\documentclass[11pt]{article}
\usepackage{amssymb}
\usepackage{acl}

\usepackage{times}
\usepackage{latexsym}

\usepackage[T1]{fontenc}

\usepackage[utf8]{inputenc}

\usepackage{microtype}

\usepackage{inconsolata}

\usepackage{graphicx}
\usepackage{listings}
\usepackage{xcolor}
\usepackage{soul}
\usepackage{booktabs}
\usepackage{tabularx}
\usepackage{caption}
\usepackage{siunitx}
\usepackage{placeins}
\usepackage{subcaption}
\usepackage{colortbl}
\usepackage{xcolor}
\usepackage{subcaption}  

\definecolor{backgroundjson}{RGB}{240, 239, 239}
\definecolor{jsonstring}{RGB}{162, 27, 12} 
\definecolor{jsonkeyword}{RGB}{24, 72, 59} 

\lstdefinelanguage{json}{
  basicstyle=\ttfamily\footnotesize, 
  breaklines=true,
  showstringspaces=false,
  morestring=[b]",
  stringstyle=\color{jsonstring},
  morecomment=[l]{//},
  morecomment=[s]{/*}{*/},
  morekeywords={true,false,null},
  sensitive=true
}

\usepackage{amsmath}

\title{Teaching Through Analogies: A Modular Pipeline \\for Educational Analogy Generation}

\author{
  Mariam Barakat \and Ekaterina Kochmar \\
  Mohamed bin Zayed University of Artificial Intelligence, UAE \\
  \texttt{\{mariam.barakat, ekaterina.kochmar\}@mbzuai.ac.ae}
}

\begin{document}
\maketitle
\begin{abstract}
Analogies help learners understand unfamiliar concepts by relating them to known concepts. Despite recent advances, large language models (LLMs) continue to struggle to generate analogies of comparable quality to those produced by humans. We present a modular pipeline for educational analogy generation, decomposing the task into four stages: source finding, sub-concept generation, explanation generation, and evaluation. Grounded in Structure Mapping Theory, the pipeline enables systematic, stage-by-stage analysis of how model choice and input configuration affect analogy quality. We evaluate 12 state-of-the-art LLMs across six model families on two datasets with structured sub-concept annotations (SCAR and ParallelPARC), alongside seven embedding models for closed-setting retrieval. Our results show that sub-concepts substantially improve explanation quality and closed setting retrieval precision but provide limited benefit in open-ended source generation. We further introduce an LLM-as-a-judge evaluation methodology and validate its scoring against human annotations from seven annotators, finding that Claude Sonnet 4.6 aligns more reliably with human rankings than with fine-grained absolute scores. Taken together, our findings reveal cross-stage interactions that isolated studies cannot capture, and highlight sub-concept grounding as a key driver of analogy quality generation.
\end{abstract}

\section{Introduction}
\begin{figure}[h]
    \centering
    \includegraphics[width=0.60\linewidth]{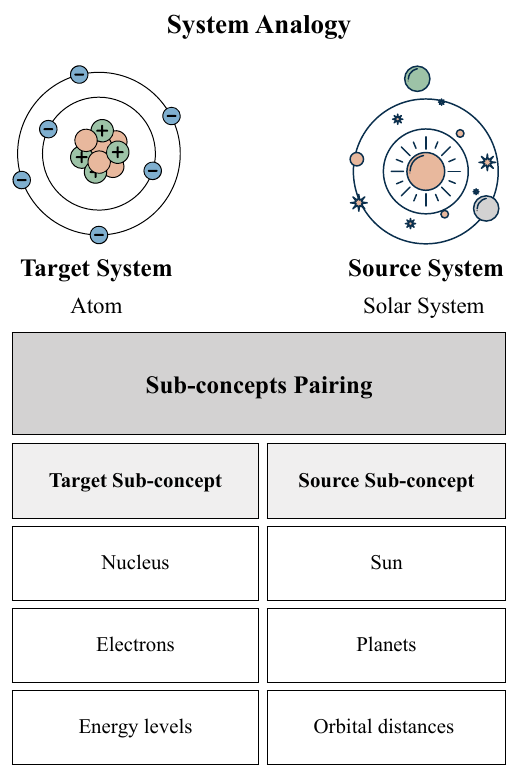}
    \caption{Illustrative example of a system analogy between the {\em atom} and the {\em solar system}, showing the source and target systems and their aligned sub-concepts.}

    \label{fig:example_diagram}
\end{figure}

As educational systems evolve in the age of AI, memorisation is no longer sufficient. Instead, students are required to be able to reason across domains and understand underlying structures -- skills that are foundational to critical thinking in a rapidly changing world. One of the effective ways to develop these abilities is through analogies \citep{eriksson2024analogy}, which facilitate learning by mapping unfamiliar knowledge (\textit{target system}) to prior knowledge (\textit{source system}) through relational structures \citep{ye2024analobench, czinczoll2022scientific, bettin2023pedagogical}. Even when a significant cognitive gap exists, intermediate ``bridging'' analogies can guide students toward comprehension \citep{brown1989overcoming}. In this work, we frame analogy generation as a modular pipeline to enable systematic analysis and improvement of each stage of the process.

Following previous research \citep{yuan2023beneath}, we use the term \textit{system} rather than \textit{concept} to emphasise that analogies map multiple components. We refer to these constituent properties as \textit{sub-concepts} throughout this paper, using "sub-'' to emphasise that they are components of the larger system rather than standalone concepts.

As shown in Figure~\ref{fig:example_diagram}, explaining the \textit{atom} (target system) through the \textit{solar system} (source system) requires the explained mapping of individual sub-concepts: the \textit{nucleus} maps to the \textit{sun}, \textit{electrons} map to \textit{planets}, and \textit{energy levels} map to \textit{orbital distances}. 

 Automatically generating high-quality analogies remains an open challenge, as poorly constructed analogies can introduce or reinforce misconceptions through incorrect mappings or oversimplification of underlying concepts \citep{zhou2025co, shao2025unlocking}.
 
 Despite recent progress, analogy generation research remains fragmented: prior work addresses retrieval, generation, or evaluation in isolation, without connecting them into a unified pipeline \citep{yuan2023beneath, yuan2023analogykb, zhai2024long, sultan2024parallelparc}. Source finding relies predominantly on embedding-based retrieval over closed candidate pools, favouring surface similarity over structural alignment \citep{yuan2023beneath, czinczoll2022scientific, jiayang2023storyanalogy}, while sub-concept annotations in SCAR \citep{yuan2023beneath} and ParallelPARC \citep{sultan2024parallelparc} are used for dataset construction rather than to guide generation. Evaluation remains under-addressed, as neither lexical metrics nor LLM-as-a-judge approaches align reliably with human judgments \citep{zhai2024long, bhavya2024long}, and most studies evaluate GPT-4 variants exclusively \citep{yuan2023beneath, kim2023metaphorian, sultan2024parallelparc, shao2025unlocking, li2024past}, leaving cross-model generalisability unexplored.

In this paper, we make three contributions: (1) a modular pipeline for educational analogy generation, grounded in Structure Mapping Theory \citep{gentner1981scientific}, decomposed into source finding, sub-concept generation, explanation generation, and evaluation; (2) a systematic cross-model evaluation across 12 LLMs from six families, finding that sub-concepts improve retrieval and explanation quality but offer limited benefit in open-ended source generation; and (3) a closed-to-open evaluation methodology, moving from controlled evaluation against gold annotations on SCAR \citep{yuan2023beneath} and ParallelPARC \citep{sultan2024parallelparc} to LLM-generated candidates with baseline, embedding-based, and re-ranking selection strategies.

Each stage in our proposed pipeline corresponds to a research question:

\noindent \textbf{RQ1: Source Finding.} Can LLMs identify suitable source systems for a given target in both closed and open settings, moving beyond predefined candidate pools? 

\noindent \textbf{RQ2: Sub-concept Generation.} Can LLMs reliably generate sub-concept mappings given a target-source pair, and at which pipeline stages do sub-concepts improve performance? 

\noindent \textbf{RQ3: Explanation Generation.} How well can LLMs produce full analogical explanations, and what input configurations yield the highest quality?

\noindent \textbf{RQ4: Evaluation.} How well does LLM-as-a-judge evaluation align with human judgments of analogy quality? Moreover, which generation model is the best at source finding in open settings?

All code and data are publicly available at: \url{https://github.com/Myriam2002/Toward-Usable-Scientific-Analogies}.

\section{Related Work}

\paragraph{Source Finding.}
Prior approaches to source finding rely primarily on embedding-based retrieval within closed candidate pools. \citet{yuan2023beneath} use greedy and Kuhn-Munkres matching but focus on mapping given concepts rather than discovering new sources. Similarly, \citet{yuan2023analogykb} employ embedding similarity, but focus only on simple one- or two-relation analogies. \citet{jiayang2023storyanalogy} show that LLMs are easily distracted by surface similarity, achieving only 30\% accuracy compared to 85\% for humans, while \citet{ye2024analobench} report substantial performance gaps on longer analogies. \citet{li2024past} further show that free generation outperforms embedding-based retrieval for analogies in the history domain, suggesting fundamental limitations of similarity-based methods. However, open-ended source discovery, where models must identify suitable sources without predefined candidates, remains largely unexplored.

\paragraph{Sub-concept Generation.}
Datasets such as SCAR \citep{yuan2023beneath} and ParallelPARC \citep{sultan2024parallelparc} provide rich sub-concept annotations linking source and target systems. However, these annotations are primarily used for dataset construction rather than as inputs to generation models. \citet{yuan2023beneath} show that sub-concept matching is critical for analogy quality, yet models achieve only 62\% accuracy compared to 86\% for humans, even when sub-concepts on both sides are provided. Moreover, evaluation is restricted to closed settings with fully specified inputs. The ability of LLMs to generate sub-concepts in open settings and their impact across pipeline stages has not been studied.

\paragraph{Explanation Generation.}
A well-matched source and accurate sub-concept mappings are necessary but not sufficient, as the explanation is what ultimately determines whether a learner grasps the analogy.
Producing explanations that capture \textit{why} an analogy holds remains particularly challenging. \citet{zhai2024long} separate concept generation and explanation generation, finding that while LLMs approach human performance on concept generation, they struggle to produce explanations with sufficient relational depth. \citet{bhavya2024long} similarly report that model-generated explanations often lack coherence or drift from the intended analogy. Overall, explanation generation remains underexplored, particularly regarding how input representations affect quality.

\paragraph{Evaluation.}
Evaluating analogy quality remains challenging. Lexical metrics such as ROUGE and BLEURT fail to capture relational soundness \citep{zhai2024long}, while more structured evaluation frameworks still fall short of fully capturing analogy quality \citep{bhavya2024long}. Human evaluation, although more reliable, is costly \citep{jiayang2023storyanalogy, czinczoll2022scientific}. LLM-as-a-judge methods offer a scalable alternative \citep{zhai2024long}, but their alignment with human judgments in analogy-specific settings remains mediocre.

\section{Methodology}

\begin{figure*}[t]
    \centering
    \includegraphics[width=\textwidth]{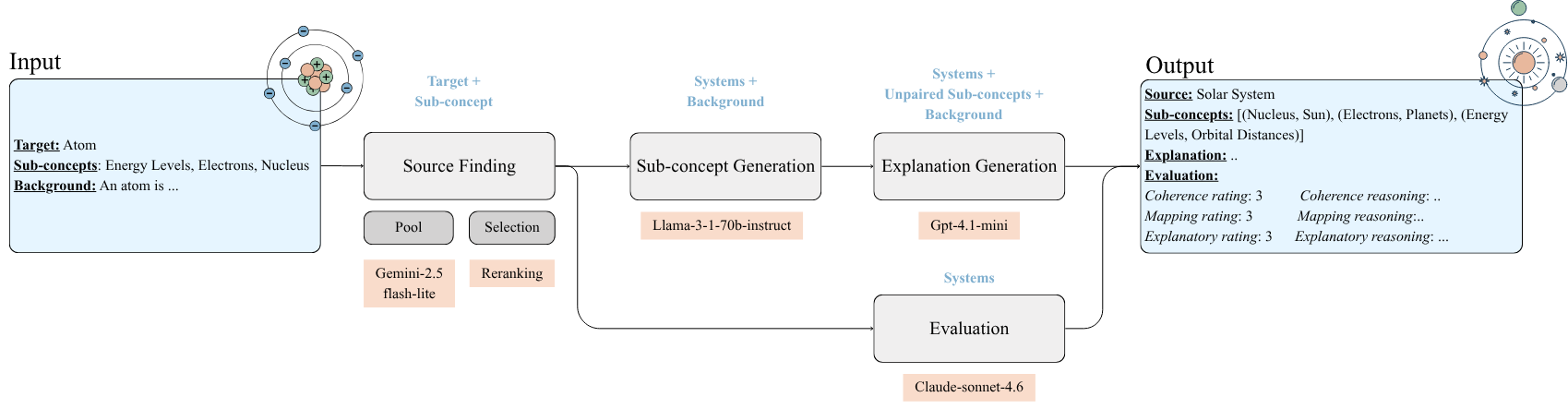}
    \caption{Overview of the proposed modular pipeline for educational analogy generation. Given a target system with its sub-concepts and background description, the framework proceeds through four stages: (1) source finding, where a pool of candidate sources is generated and selected; (2) sub-concept generation; (3) explanation generation; and (4) evaluation using an external LLM-as-a-judge. Each stage operates under input configurations tailored to its requirements, as determined by our empirical results. The output illustrates the final generated analogy, including aligned sub-concepts, explanation, and evaluation scores.}
    \label{fig:methodology_rq3}
\end{figure*}

We present a systematic evaluation of the proposed modular pipeline for educational analogy generation, analysing how model choice and input configuration influence performance at each stage. The pipeline is decomposed into four stages aligned with our research questions: source finding (RQ1), sub-concept generation (RQ2), explanation generation (RQ3), and evaluation (RQ4). Figure~\ref{fig:methodology_rq3} provides an overview of the experimental setup. To illustrate the full pipeline, consider the example in Figure \ref{fig:example_diagram}, where the target system \textit{atom} is explained using the source \textit{solar system}. In the source-finding stage, candidate source systems are retrieved based on their similarity to the target. Next, sub-concept generation aligns components such as {\em nucleus} $\rightarrow$ {\em sun} and {\em electrons} $\rightarrow$ {\em planets}. The explanation generation stage then produces a natural-language explanation linking these correspondences. Finally, the analogy is evaluated based on coherence, mapping soundness, and explanatory power. This example demonstrates how each stage contributes to the final analogy.

\subsection{Datasets}
\label{sec:datasets}
We use two datasets with explicit sub-concept annotations, enabling controlled evaluation across pipeline stages.\footnote{Throughout the paper, we adopt the field names from SCAR for clarity, and align corresponding fields from other datasets to match this structure.}

\textbf{SCAR} \citep{yuan2023beneath} comprises 400 system-level analogy instances across 13 scientific domains, created by human annotators. Each instance pairs a target system with a source system and is annotated with background descriptions, multiple sub-concept mappings, and explanations for these mappings (1,614 total annotated mappings). A representative example is provided in Appendix~\ref{app:scar}.

\textbf{ParallelPARC} \citep{sultan2024parallelparc} extends the ProPara dataset \citep{tandon2018reasoning} by transforming scientific-process descriptions into analogy pairs through a pipeline of LLM generation, automatic labelling, and human validation. It introduces 310 gold analogies, each annotated with relational alignments between corresponding properties in the format \texttt{(entity, verb, entity) mapped to (entity, verb, entity)}. Please refer to Appendix~\ref{app:parallel} for an example.

\subsection{Models}
\label{sec:models}

\paragraph{Closed Setting.}
For source retrieval in the closed setting, we evaluate a diverse set of embedding models: OpenAI's \texttt{embedding-small} and \texttt{embedding-large},\footnote{\url{https://platform.openai.com/docs/guides/embeddings}} \texttt{E5-large},\footnote{\url{https://huggingface.co/intfloat/e5-large}} \texttt{BGE-large},\footnote{\url{https://huggingface.co/BAAI/bge-large-en}} \texttt{MPNet-base},\footnote{\url{https://huggingface.co/sentence-transformers/all-mpnet-base-v2}} \texttt{MiniLM-L6},\footnote{\url{https://huggingface.co/sentence-transformers/all-MiniLM-L6-v2}} and \texttt{SPECTER}.\footnote{\url{https://huggingface.co/allenai/specter}} These models cover complementary paradigms: proprietary general-purpose embeddings (OpenAI), open-source retrieval-oriented models (\texttt{E5-large}, \texttt{BGE-large}), sentence-similarity baselines (\texttt{MPNet-base}, \texttt{MiniLM-L6}), and a domain-specialised scientific encoder (\texttt{SPECTER}), enabling comparison across models that differ in scale, training objective, and domain specificity.

\paragraph{Open Setting.} For open-ended generation across all three stages, we use the \texttt{DSPy} framework \citep{khattab2022demonstrate, khattab2024dspy} for consistent input/output interfaces. We evaluate 12 LLMs across six families: OpenAI's \texttt{GPT-4.1} (mini, nano), \texttt{GPT-OSS} (20B, 120B), Meta's \texttt{LLaMA~3.1} (8B, 70B, 405B) Instruct, Google's \texttt{Gemini~2.5~Flash-Lite}, xAI's \texttt{Grok-4-Fast}, \texttt{DeepSeek-R1}, and Alibaba's \texttt{Qwen3} (14B, 32B). Models were selected based on OpenRouter's public rankings in academia and science,\footnote{\url{https://openrouter.ai/rankings}} balancing benchmark performance and cost-efficiency. All models were run at temperature 0.2 for consistency. Models were accessed via OpenAI, OpenRouter, and DeepInfra APIs.\footnote{OpenAI: \url{https://openai.com}, OpenRouter: \url{https://openrouter.ai}, DeepInfra: \url{https://deepinfra.com}}

\subsection{Source Finding (RQ1): Pool $\&$ Selection}
\paragraph{Closed Setting: Embedding-based Retrieval.}
\label{sec:closed-retrieval}

Following prior work \citep{czinczoll2022scientific, yuan2023beneath}, we embed all unique source systems from SCAR and ParallelPARC using the seven models from Section~\ref{sec:models} and retrieve the top-20 candidates per target via cosine similarity. We test four input configurations with increasing context: (1) \textbf{target only} to assess surface-level retrieval; (2) \textbf{+background} to test whether definitional context helps; (3) \textbf{+sub-concepts} to assess the role of structural alignment; and (4) \textbf{sub-concepts \& background} to evaluate the effect of combining both.

To go beyond similarity-based ranking, we introduce an LLM reranking stage: the top-10 candidates from the best-performing embedding configuration are passed to \texttt{LLaMA-3.1-70B-Instruct}, which selects the top-3 (Appendix~\ref{app:analogypick3}). We evaluate five reranking configurations: (1) \textbf{target only} as a baseline; (2) \textbf{+background} to test definitional context; (3) \textbf{+gold sub-concepts} to assess structural alignment with gold annotations; (4) \textbf{+sub-concepts \& background} to combine both signals; and (5) \textbf{+generated sub-concepts} to test whether generated sub-concepts can substitute for gold annotations in downstream selection (see Subsection \ref{ssec:sub-concept}). \label{sec:reranker} 

We evaluate retrieval using \textit{Hit@K} (Eq.~\ref{eq:hit_k}), which measures whether at least one gold source appears in the top-$K$ results. This evaluation directly measures the quality of the retrieval component by assessing whether embedding models successfully retrieve appropriate source systems for a given target.

\begin{equation}
  \label{eq:hit_k}
  \text{Hit@K} = \frac{1}{N} \sum_{i=1}^{N} \mathbb{I}(\text{rank}_i \leq K)
\end{equation}

\paragraph{Open Setting: LLM-based Generation.}
\label{sec:open-generation}

In the open setting, each of the 12 LLMs generates 20 ranked candidate sources from scratch, evaluated on SCAR only due to its educational relevance and computational constraints. Following closed-setting results, we test two input configurations: (1) \textbf{target only} and (2) \textbf{+sub-concepts}.

Since multiple candidates are generated, we evaluate three selection strategies: (1) \textbf{baseline top-1}, selecting the first output from an LLM; (2) \textbf{embedding-based top-1}, selecting the candidate with the highest cosine similarity to the target using \texttt{text-embedding-3-small}, strong in closed-setting retrieval and cost-efficient; and (3) \textbf{re-ranker top-1}, applying the LLM-based reranking framework from Subsection ~\ref{sec:reranker} with the \textbf{name $\&$ subconcepts} configuration, selected based on its strong performance in the closed setting.

We report both {\em exact Hit@K} (Eq.~\ref{eq:hit_k}) and \textit{semantic Hit@K}. A generated source is considered correct if its embedding similarity to any gold source exceeds 0.374, corresponding to the upper tertile of pairwise gold similarities in SCAR (see Appendix~\ref{app:threshold} for more details). To avoid evaluation bias, we use \texttt{all-MiniLM-L6-v2} for semantic Hit@K scoring rather than \texttt{text-embedding-3-small}, which is used in the embedding-based Top-1 selection step. Using the same model for both selection and evaluation might unfairly favour that configuration over others.

\subsection{Sub-concept Generation (RQ2)}
\label{ssec:sub-concept}

\begin{figure}[h]
    \centering
    \includegraphics[width=0.48\textwidth]{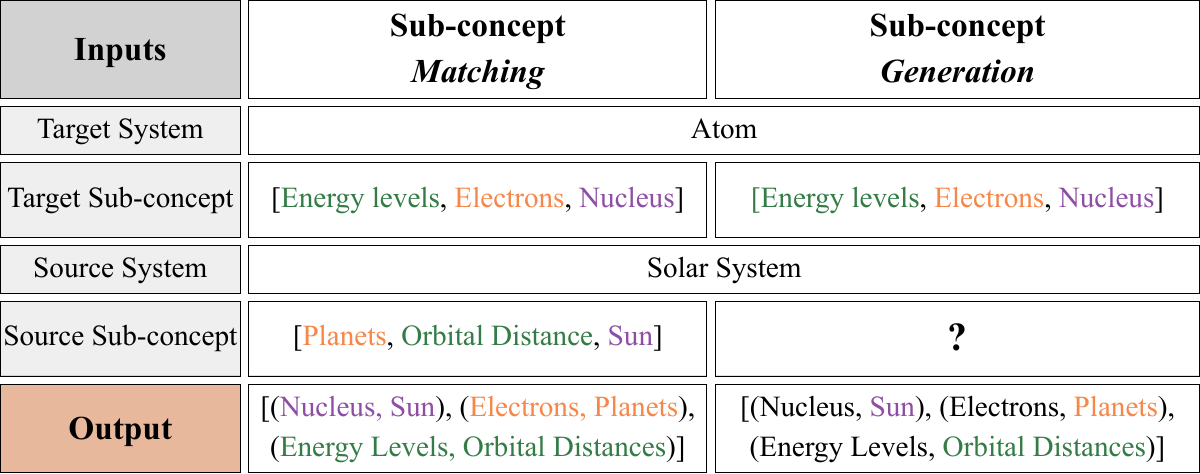}
    \caption{Sub-concept matching and generation stage (RQ2). 
    Given a target system (e.g., {\em atom}) and a source system (e.g., {\em solar system}), the model performs sub-concept matching using gold sub-concepts, and sub-concept generation when source sub-concepts are unavailable, producing aligned sub-concept pairs.}
    \label{fig:methodology_rq12}
\end{figure}

Before tackling the harder generation task, we revisit sub-concept matching -- where models align given sub-concepts rather than produce new ones -- to establish how well current LLMs capture structural correspondences and how far the field has advanced since \citet{yuan2023beneath}. Figure~\ref{fig:methodology_rq12} highlights the distinction between the two tasks.

\paragraph{Sub-concept Matching.}
We re-evaluate sub-concept matching on SCAR, using its reported system accuracy (62\% for GPT-4 vs.\ 86\% for human annotators) as a reference point. Models receive randomly shuffled lists of target and source sub-concepts and must identify the correct correspondences. Then, we test two scenarios: (1) \textbf{without background}, where models rely only on sub-concept labels to recognise structural alignments; and (2) \textbf{with background} (see Listing \ref{lst:scar-json} in Appendix \ref{app:scar}). We compare performance across the 12 LLMs to determine if the gap between model and human performance can be narrowed. For evaluation, we use \textit{System Accuracy} \citep{yuan2023beneath}, which assigns 1 if all pairs are correct and 0 otherwise.

\paragraph{Sub-concept Generation.}

To test whether LLMs can generate source sub-concepts in open settings, we test two configurations: given the target system, its sub-concepts, and the source system name, models must produce the corresponding source sub-concepts either (1) \textbf{without background} or (2) \textbf{with background}.

To evaluate, we use \textit{Semantic Match Accuracy (SMA)}. It uses cosine similarity between embeddings built with {\tt all-MiniLM-L6-v2}.\footnote{{\tt all-MiniLM-L6-v2}:\url{https://huggingface.co/sentence-transformers/all-MiniLM-L6-v2}} A fixed 0.7 threshold is adopted as a conservative heuristic for MiniLM embeddings, balancing semantic sensitivity with false-positive robustness.

\begin{equation}
  \label{eq:semantic_match}
  \text{SMA} = \frac{\text{count}(\text{similarity} \ge 0.7)}{\text{total concepts}}
\end{equation}

\subsection{Explanation Generation (RQ3)}
\label{sec:explanation-generation}
The third stage evaluates whether LLMs can articulate why an analogy holds, given a target and source. Each model generates a natural-language explanation. To isolate the contribution of different input signals, we test six configurations: (S1)~\textbf{names only}, (S2)~\textbf{+background}, (S3)~\textbf{+unpaired sub-concepts}, (S4)~\textbf{+unpaired sub-concepts \& background}, (S5)~\textbf{+paired mappings}, and (S6)~\textbf{+paired mappings \& background} (see the prompts in Appendix~\ref{app:explanation-prompt}). Throughout the results, we refer to these configurations as S1--S6 for brevity.

We evaluate generated explanations against gold references using cosine similarity over \texttt{all-mpnet-base-v2} \citep{reimers-2019-sentence-bert} embeddings known for their ability to compare sentences to each other, capturing semantic equivalence beyond surface variation, and then we compare performance across input settings.

\subsection{Evaluation (RQ4)}
\paragraph{LLM-as-a-judge.}
\label{sec:llm-judge}
To close the pipeline, we turn to evaluation -- employing \texttt{Claude Sonnet 4.6}\footnote{\url{https://openrouter.ai/anthropic/claude-sonnet-4.6}} as an external LLM judge, independent of all 12 pipeline models to reduce evaluation bias. Analogies are scored on three dimensions (using 1--3 scale): \textbf{coherence} (does the pairing make intuitive sense?), \textbf{mapping soundness} (are source--target correspondences structurally valid?), and \textbf{explanatory power} (does the analogy help a learner understand the target?). We use few-shot prompting with 10 calibrated examples spanning the full score range, as preliminary experiments without examples yielded less stable judgments. The full prompt is provided in Appendix~\ref{app:llm-as-judge}. We apply this judge across all 12 models to identify which one produces the highest-quality analogies overall.

\paragraph{Human Evaluation.}
\label{sec:human-eval}

To validate the LLM-based ratings, we collect human judgments from seven graduate students in AI and Business.\footnote{Annotation form: \url{https://myriam2002.github.io/analogy-annotation/}} Prior to annotation, participants were provided with detailed instructions describing the task, including definitions of the three evaluation dimensions (coherence, mapping soundness, and explanatory power), along with illustrative examples to calibrate their understanding. Annotators evaluate 45 analogy pairs -- 15 target concepts across five domains (check Appendix \ref{app:45table} for all analogies tested), each paired with three candidate sources -- scoring each on the same three dimensions used by the LLM judge. In addition to scoring individual dimensions, annotators rank candidate analogies based on overall learning usefulness, capturing whether the analogy is intuitive and helpful for understanding the target concept. This ranking provides a qualitative, learner-oriented perspective that complements the dimension-level scores, yielding a total of 165 annotated datapoints.\footnote{Confidence scores were collected for ranking but not used in the current analysis; incorporating them to weigh annotator reliability is left for future work.}

We measure inter-annotator reliability using Krippendorff's $\alpha$ \citep{krippendorff2011computing} for ordinal ratings, Kendall's $W$ \citep{kendall1990rank} for ranking concordance across annotators, and Spearman's $\rho$ to compare human scores against LLM judge scores.

\section{Results and Analysis}

\subsection{Source Finding (RQ1)}
\paragraph{Closed Setting.}
Embedding-based retrieval shows consistent but dataset-dependent trends. On SCAR, adding sub-concepts yields the largest gains: {\tt OpenAI-Large} improves from 0.710 to 0.792 and {\tt OpenAI-Small} from 0.677 to 0.780, with the average increase from 0.557 to 0.673. Adding background provides no benefit (avg.\ 0.629), likely because sub-concepts already capture structural alignment, and extra context introduces noise. On ParallelPARC, performance is near-saturated across settings (name-only avg.\ 0.876), with sub-concepts + background performing best ({\tt OpenAI-Large}: 0.942, {\tt OpenAI-Small}: 0.932, avg.\ 0.908), as its backgrounds provide richer relational signal than SCAR descriptions. Across both datasets, {\tt OpenAI} embeddings consistently lead, while {\tt SPECTER} lags (SCAR avg.\ 0.437; ParallelPARC avg.\ 0.781), reflecting weaker alignment with the task. Full results are in Appendix~\ref{app:retrieval_results}.

For selection, we compare five configurations: re-ranking with names only, background, gold sub-concepts, gold sub-concepts + background, generated sub-concepts, and embedding-only retrieval. Re-ranking with gold sub-concepts achieves the best Hit@1 (24.5\%), followed by background (22.2\%), gold + background (20.8\%), embedding-only (19.5\%), generated sub-concepts (18.2\%), and names only (17.5\%). At Hit@2 and Hit@3, embedding-only performs best (30.5\%, 40.0\%), while gold sub-concepts rank third (31.8\%, 35.8\%), indicating that embeddings better retrieve candidates, whereas LLM re-ranking excels at top-1 selection. Generated sub-concepts outperform the name-only baseline (18.2\% vs \ 17.5\% at Hit@1), showing that the pipeline produces meaningful structure without gold annotations, though a gap remains. Overall, embeddings narrow the search space, while LLM reasoning over sub-concepts is key for precise selection.

\paragraph{Open Setting.}

Adding sub-concepts to the generation prompt yields only marginal gains (avg.\ 0.4-1.6 pp across all K and models), suggesting limited benefit compared to their substantial impact in retrieval, while incurring additional prompt cost, making target-only prompting the more practical default. The two exceptions are {\tt LLaMA-405B} and {\tt Qwen3-32B}, which show a slight decrease when sub-concepts are added. In absolute terms, the strongest Hit@20 results are achieved under the target with sub-concepts mode, with {\tt Gemini-2.5} (75.4\%) and {\tt GPT-OSS-120B} (75.3\%) leading. However, top models cluster within a few percentage points of each other, suggesting that scale alone does not determine generation quality. Per-model results are in Appendix~\ref{app:targetvssubconcept}.

\subsection{Sub-concept Generation (RQ2)}

\definecolor{lightgray}{gray}{0.88} 

\begin{table}[t]
\centering
\small
\begin{tabular}{lcccc}
\toprule
& \multicolumn{2}{c}{\textbf{Matching}} & \multicolumn{2}{c}{\textbf{Generation}} \\
\cmidrule(lr){2-3} \cmidrule(lr){4-5}
\textbf{Model} & \textbf{+BG} & \textbf{--BG} & \textbf{+BG} & \textbf{--BG} \\
\midrule
Grok-4       & 0.76 & \cellcolor{lightgray}\textbf{0.78} & \cellcolor{lightgray}0.42 & 0.31 \\
DeepSeek-R1  & 0.70 & \cellcolor{lightgray}0.76 & \cellcolor{lightgray}0.45 & 0.33 \\
GPT-OSS-120B & 0.69 & \cellcolor{lightgray}0.72 & \cellcolor{lightgray}0.34 & 0.27 \\
Qwen3-14B    & 0.68 & \cellcolor{lightgray}0.70 & \cellcolor{lightgray}0.42 & 0.31 \\
GPT-OSS-20B  & 0.66 & \cellcolor{lightgray}0.68 & \cellcolor{lightgray}0.39 & 0.28 \\
GPT-4.1-mini & 0.66 & \cellcolor{lightgray}0.68 & \cellcolor{lightgray}0.42 & 0.31 \\
Qwen3-32B    & 0.65 & \cellcolor{lightgray}0.71 & \cellcolor{lightgray}0.40 & 0.31 \\
Llama-70B    & 0.57 & \cellcolor{lightgray}0.60 & \cellcolor{lightgray}\textbf{0.47} & 0.32 \\
Llama-405B   & 0.56 & \cellcolor{lightgray}0.62 & \cellcolor{lightgray}\textbf{0.47} & 0.33 \\
Gemini-2.5   & 0.54 & \cellcolor{lightgray}0.64 & \cellcolor{lightgray}0.42 & 0.29 \\
Llama-8B     & 0.40 & \cellcolor{lightgray}0.46 & \cellcolor{lightgray}0.44 & 0.27 \\
GPT-4.1-nano & 0.37 & \cellcolor{lightgray}0.46 & \cellcolor{lightgray}0.34 & 0.27 \\
\bottomrule
\end{tabular}
\caption{System accuracy (Matching) and semantic match accuracy (Generation) with (+BG) and without (--BG) background descriptions. Bold indicates the best score per column. Gray shading indicates the higher-scoring mode per task for each model. Models are sorted by matching accuracy.}
\label{tab:matching_generation}
\end{table}

Table~\ref{tab:matching_generation} contrasts sub-concept matching and generation across models. For matching, performance is slightly higher \emph{without} background descriptions (avg.\ 0.650 vs.\ 0.603), suggesting additional context hinders alignment. {\tt Grok-4} leads at 0.78, surpassing the prior state-of-the-art of 0.62~\cite{yuan2023beneath}, with gains attributed to model choice and prompt design (Appendix~\ref{app:ablation}). {\tt GPT-4.1-nano} performs worst (0.37) despite not being the smallest model. These results motivate the harder generation setting explored next.

For generation, background descriptions provide a clear benefit, raising average semantic match accuracy from 0.298 to 0.412. The LLaMA family leads, with {\tt LLaMA-3.1-70B} and {\tt LLaMA-3.1-405B} both reaching 0.47, while GPT-based variants consistently occupy the lower half. Notably, {\tt GPT-4.1-nano} and {\tt GPT-OSS-120B} (0.34) underperform the smaller {\tt GPT-OSS-20B} (0.39), confirming that scale alone does not guarantee stronger generation.

Based on these findings, we adopt {\tt LLaMA-3.1-70B} with background descriptions for all subsequent generation experiments.

\subsection{Explanation generation (RQ3)}
 The most apparent pattern is the large jump in similarity when sub-concepts are introduced: average scores rise from 0.724 in S1--S2 to 0.854 with unpaired sub-concepts (S3--S4) and 0.830 with paired sub-concepts (S5--S6). By contrast, adding background descriptions has a negligible effect across all three setting pairs (at most $+$0.006), consistent with our earlier matching findings. The best setting overall is S5 (systems + paired sub-concepts).

{\tt GPT-4.1-Mini} is the strongest model overall (mean 0.827) and is the best performer in four of the six settings (S3--S6), indicating strong generalisation across sub-concept conditions. In the system-only settings (S1--S2), where scores are uniformly low and differences between models are small (range $\approx$  0.03), no single model stands out, with {\tt GPT-OSS-120B} and {\tt Llama-70B} marginally leading in S1 and S2, respectively. Check Appendix \ref{app:explanation_heat} for the detailed heatmap.

\subsection{Evaluation (RQ4)}
\paragraph{LLM-as-a-judge.}
\begin{figure}[h]
    \centering
    \includegraphics[width=\linewidth]{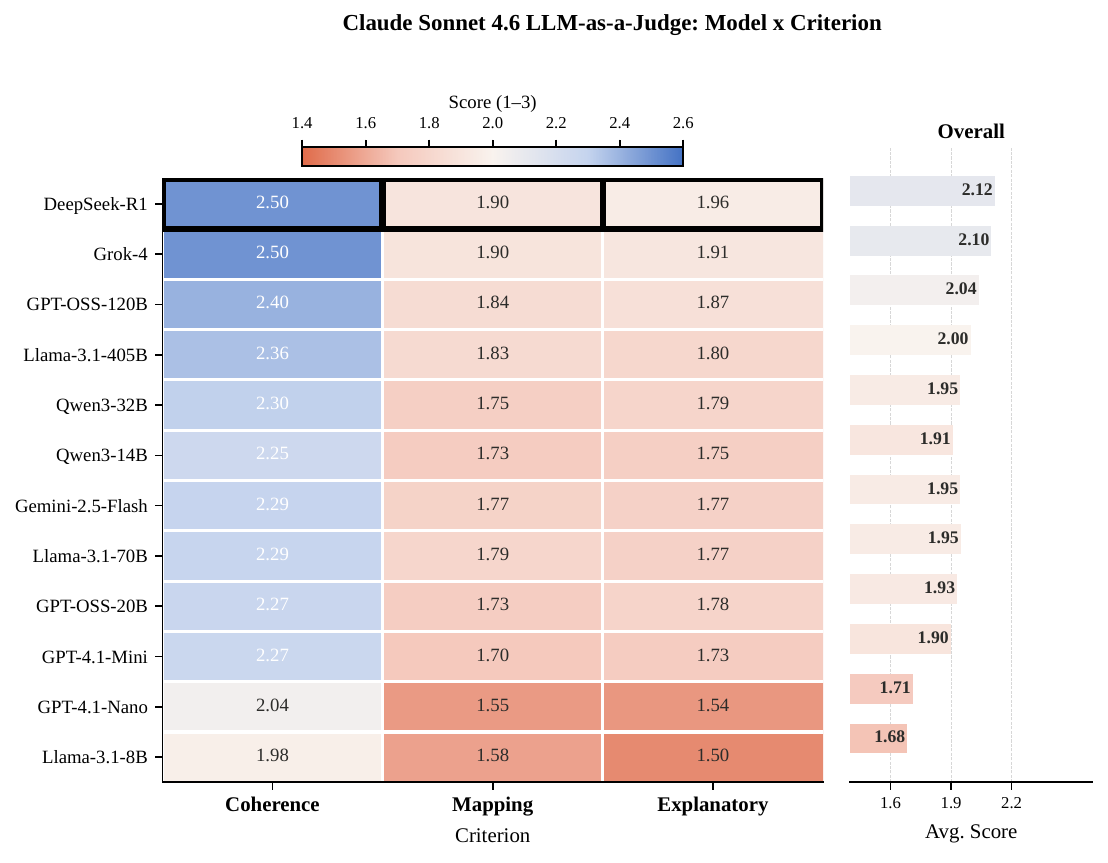}
    \caption{Per-criterion scores assigned by Claude Sonnet 4.6 acting as an LLM judge. Scores are on a 1--3 scale across three criteria: Coherence, Mapping, and Explanatory power. Black borders indicate the top-performing model per criterion. }
    \label{fig:judge_leaderboard}
\end{figure}
Figure~\ref{fig:judge_leaderboard} shows LLM-as-a-judge scores across generation models. \texttt{DeepSeek-R1} and \texttt{Grok-4} achieve the highest overall performance (2.12 and 2.10), while \texttt{GPT-4.1-Nano} and \texttt{LLaMA-3.1-8B} perform the worst (1.71 and 1.68). The remaining models cluster closely in the middle, indicating relatively small differences across most systems. Across all models, coherence scores are consistently higher than mapping and explanatory scores, suggesting that models often produce intuitively plausible analogies even when structural mappings are incomplete or explanations are less informative.

\begin{table}[h]
\centering
\resizebox{\columnwidth}{!}{%
\begin{tabular}{lcccccc}
\hline
\textbf{Dimension} &
\multicolumn{2}{c}{\textbf{Baseline}} &
\multicolumn{2}{c}{\textbf{Embeddings}} &
\multicolumn{2}{c}{\textbf{Reranker}} \\
 & mean & std & mean & std & mean & std \\
\hline
Coherence   & \textbf{2.463} & 0.652 & 1.956 & 0.748 & 2.437 & 0.677 \\
Mapping     & 1.839 & 0.560 & 1.550 & 0.583 & \textbf{1.878} & 0.561 \\
Explanatory & \textbf{1.902} & 0.622 & 1.484 & 0.581 & 1.894 & 0.640 \\
Avg. Score  & 2.068 & 0.541 & 1.663 & 0.575 & \textbf{2.069} & 0.550 \\
\hline
\end{tabular}%
}
\caption{Comparison of analogy quality across selection strategies using LLM-as-a-judge scores (1--3 scale).}
\label{tab:llm_judge_stats}
\end{table}

Table~\ref{tab:llm_judge_stats} compares selection strategies under LLM-as-a-judge evaluation. The baseline and re-ranking approaches achieve close average scores (2.068 vs.\ 2.069), but differ in their strengths: the baseline yields slightly higher coherence, while re-ranking improves mapping quality. This indicates that LLM-based selection refines structural alignment without substantially affecting overall performance.

\paragraph{Human Evaluation.}

Inter-annotator agreement is measured using Krippendorff’s $\alpha$. Agreement is fair for \textit{coherence} (0.382) and \textit{explanatory power} (0.381), and lower for \textit{mapping soundness} (0.342) -- the most technically demanding dimension. Agreement is highest for the overall ranking (0.479) -- where annotators rank the three candidate sources per target by overall learning usefulness -- reflecting that relative ordering is easier to agree on than assigning consistent absolute scores across all three dimensions; for instance, two annotators may agree that source A is best, yet still differ on whether to rate its coherence a 2 or a 3. Appendix~\ref{app:human_agreement} provides detailed ranking statistics.

Spearman correlation ($\rho$) between Claude and human annotators varies across both annotators and dimensions. Agreement is highest for \textit{explanatory power} (mean $\rho = 0.511$, range $0.269$--$0.678$) and lowest for \textit{mapping soundness} (mean $\rho = 0.342$, range $0.124$--$0.507$), consistent with LLM-as-a-judge trends. Coherence falls between these extremes (mean $\rho = 0.368$, range $0.130$--$0.581$), reflecting moderate agreement on whether analogies are intuitively meaningful. Annotator-level variability is also notable: the weakest alignment consistently comes from the annotator whose academic background differed most from the rest of the group, which may be partially explained by the domain composition of the evaluation set, where five of the fifteen target concepts were drawn from the computer science domain.

\begin{table}[h]
\centering
\small
\begin{tabular}{lccc}
\hline
\textbf{Annotator} & $\rho$ & \textbf{p-value} & \textbf{Significant} \\
\hline
ANN-6JJP46 & 0.474 & 0.0010 & True \\
ANN-6LG7WY & 0.705 & 0.0000 & True \\
ANN-DUNVTH & 0.307 & 0.0404 & True \\
ANN-H2NTMS & 0.694 & 0.0000 & True \\
ANN-KKN85E & 0.463 & 0.0014 & True \\
ANN-Q777H6 & 0.128 & 0.4012 & False \\
ANN-VM9J6N & 0.694 & 0.0000 & True \\
\hline
\end{tabular}
\caption{Spearman correlation between Claude’s implicit ranking and human rankings.}
\label{tab:spearman_ranking}
\end{table}

Table~\ref{tab:spearman_ranking} compares Claude’s implicit ranking of analogies with human rankings. In contrast to the dimension-level results, ranking agreement is generally stronger, with most annotators showing moderate-to-high correlation ($\rho$ between 0.46 and 0.70) and statistically significant results (p < 0.05). The strongest alignment is observed for ANN-6LG7WY ($\rho = 0.705$), while ANN-Q777H6 again shows no significant agreement. These results indicate that Claude more reliably captures relative preferences between analogies than assigning consistent absolute scores, suggesting that LLM-as-a-judge is better suited for comparative evaluation settings than for fine-grained scoring.

\begin{figure}[h]
    \centering
    \includegraphics[width=\linewidth]{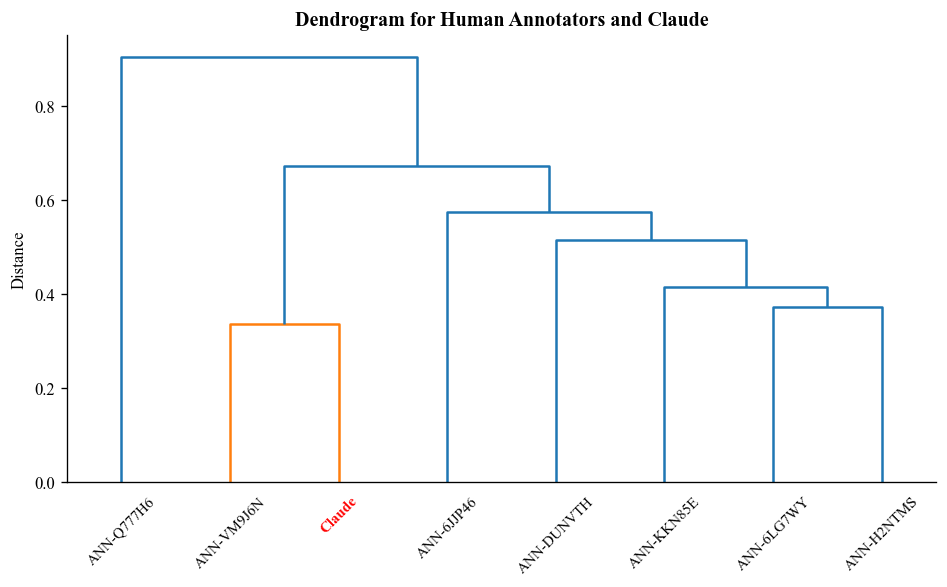}
    \caption{Hierarchical clustering of human annotators and Claude based on their scoring patterns across evaluation dimensions. Distances reflect dissimilarity in judgments (computed from Spearman correlations). Claude clusters closely with a subset of annotators (e.g., ANN-VM9J6N), while other annotators (e.g., ANN-Q777H6) form more distant branches.}
    \label{fig:dendrogram}
\end{figure}

Figure~\ref{fig:dendrogram} further examines agreement patterns through hierarchical clustering of annotators and Claude. The dendrogram reveals that Claude groups closely with a subset of annotators, suggesting similar evaluation behavior, while others form more distant clusters, indicating systematic differences in scoring. In particular, annotators with lower correlation scores (e.g., ANN-Q777H6) appear as outliers, consistent with the results in Table ~\ref{tab:spearman_ranking}. This analysis reinforces that, although Claude aligns well with the majority of annotators, human judgments themselves are not fully consistent, highlighting the inherent subjectivity of analogy evaluation.

\section{Conclusion}
We presented a modular pipeline for educational analogy generation decomposed into source finding, sub-concept generation, explanation generation, and evaluation, systematically evaluated across 12 LLMs and 7 embedding models on SCAR and ParallelPARC. A central finding is that the best model and configuration differ across stages: {\tt Grok-4-Fast} leads sub-concept matching without background descriptions, while {\tt LLaMA-3.1-70B} with background is best for generation; {\tt GPT-4.1-Mini} with unpaired sub-concept mappings achieves the strongest explanation quality; and OpenAI large embeddings with sub-concepts lead closed-setting retrieval, whereas the target-only configuration is more practical in the open setting. Sub-concept grounding consistently improves retrieval precision and explanation quality but provides limited benefit in open-ended source generation. For evaluation, {\tt DeepSeek-R1} and {\tt Grok-4-Fast} produce the highest-quality analogies, and \texttt{Claude Sonnet 4.6} aligns more reliably with human rankings than absolute scores, suggesting LLM-as-a-judge is better suited to comparative settings. Taken together, these findings demonstrate that no single model dominates across all stages, reinforcing the necessity of a modular, stage-aware approach to analogy generation.

\section*{Limitations}

Several limitations of this work warrant acknowledgment. First, all experiments are conducted exclusively on English-language data, leaving the pipeline’s behaviour on morphologically richer or structurally different languages unexplored. In such settings, both embedding-based retrieval and generation quality may differ, and evaluation metrics such as Semantic Match Accuracy (SMA) may require recalibration.

Second, the 0.7 cosine similarity threshold used for SMA is a fixed heuristic selected for \texttt{MiniLM-L6} embeddings rather than empirically validated. Its suitability for more abstract or diverse sub-concepts remains unclear, and alternative thresholds could meaningfully affect generation results.

Third, while each pipeline stage is evaluated independently, we do not evaluate the pipeline end-to-end. In a full pipeline setting, errors are likely to accumulate and propagate across stages. For example, an incorrect source selected during the retrieval stage would lead to flawed sub-concept mappings and explanations downstream. This makes source finding a particularly critical stage, as early errors can disproportionately affect final analogy quality. Future work should investigate strategies to mitigate such compounding effects, such as iterative refinement, confidence-aware filtering, or alternative candidate selection mechanisms.

Fourth, the current architecture relies primarily on dense embedding models for retrieval and does not explore hybrid search strategies. Combining dense and sparse retrieval methods, or incorporating re-ranking models more extensively, could improve the robustness and quality of source selection, particularly in open-ended settings.

Fifth, analogy quality is inherently subjective: an analogy that is effective for one learner may not be equally meaningful for another. While this subjectivity is partially reflected in moderate inter-annotator agreement, our evaluation does not explicitly model user-dependent preferences. This limits the ability to assess how well generated analogies generalise across diverse learner populations. Future work should investigate how to construct learner profiles and incorporate them into the pipeline to generate personalised analogies tailored to factors such as cultural background, age, prior knowledge, and individual interests or hobbies.

Sixth, all pipeline stages are evaluated in a single-pass setting. We do not measure output variance across runs, which means that reported results reflect point estimates rather than stable averages. Although a low temperature (0.2) is used, some variability is expected, particularly for smaller models. Moreover, since ParallelPARC was used in only one stage of the pipeline (embedding-based retrieval) and the remaining stages relied solely on SCAR, the generalizability of the results across datasets may be limited.

Finally, the human evaluation study involves six out of seven graduate-level annotators from a single institution, which may introduce demographic and disciplinary homogeneity. Inter-annotator agreement is moderate, reflecting the inherent subjectivity of analogy evaluation and suggesting that a larger, more diverse annotator pool could yield different conclusions. In addition, open-setting source finding is evaluated only on SCAR due to computational constraints, limiting the generalisability of these findings to other domains of different difficulty.

\section*{Ethical Considerations}

We do not anticipate any ethical risks related to this study.

\bibliography{custom}

\onecolumn
\appendix

\section{Detailed Record of Datasets}

Listing \ref{lst:scar-json} and \ref{lst:parc} shows a representative JSON example of the two datasets.
\label{app:scar}
\begin{lstlisting}[caption={Representative JSON example from SCAR},label={lst:scar-json}]
{
  "id": 3,
  "lang": "en",
  "system_a": "Respiratory system",
  "system_b": "Engine",
  "mappings": [
    ["oxygen", "fuel"],
    ["lungs", "combustion chamber"],
    ["breathing muscles", "piston"]
  ],
  "system_a_domain": "Biology",
  "system_b_domain": "Physics",
  "system_a_background": "The respiratory system ...",
  "system_b_background": "An engine or motor ...",
  "Explanation": [
    "Oxygen corresponds to fuel: ...",
    "Lungs correspond to the combustion chamber: ...",
    "Breathing muscles correspond to the piston: ..."
  ]
}
\end{lstlisting}

\label{app:parallel}
\begin{lstlisting}[ caption={Representative JSON from ParallelPARC}, label={lst:parc}]
{
  "sample_id": 1,
  "source_subject": "What causes a volcano to erupt?",
  "source_domain": "Natural Sciences",
  "target_domain": "Engineering",
  "target_subject": "What causes a boiler to explode?",
  "target_field": "Chemical Engineering",
  "relations": [
    "(magma, heats, underground water) like (steam, heats, liquid)",
    "(pressure, builds, inside the volcano) like (pressure, builds, inside the boiler)",
    "(magma, pushes, against the walls of the volcano) like (steam, pushes, against the walls of the boiler)"
  ],
  "source_paragraph": "When magma heats up underground water, pressure begins to build up inside the volcano, leading to an eruption.",
  "target_paragraph": "Steam heats the liquid inside the boiler, causing the pressure to build up until the boiler can no longer contain it.",
  "analogy_type": "far analogy",
  "sum_vote_analogy": 3.0
}
\end{lstlisting}

\section{Prompt for LLM Re-ranker}
\label{app:analogypick3}
\begin{lstlisting}[caption={Prompt used to select the top-3 analogies from the top-10 potential sources. This was wrapped in DSPY function as instruction, alongside specific input and output variables.}]
You are selecting the best analogous source concepts for a scientific analogy.

Your task:
1. Analyze the target concept and its properties
2. Review each candidate source and its generated analogous properties
3. Select the 3 candidates whose properties BEST map to the target properties

Selection criteria:
- Strong structural/functional correspondence between source and target properties
- The source concept should be familiar and help explain the unfamiliar target
- Prefer sources with clear, well-mapped properties over vague ones

Return the EXACT names of your top 3 selected candidates.
"""
\end{lstlisting}

\section{Threshold Decision for Semantic Hit@K}
\label{app:threshold}

To define a threshold for semantic Hit@K, we analyse the distribution of source-to-source similarity scores in the SCAR dataset. Specifically, we compare similarity values between \textit{correct} source pairs (i.e., gold analogies) and \textit{incorrect} pairs obtained by randomly shuffling sources.

\begin{figure}[h]
    \centering
    \includegraphics[width=\linewidth]{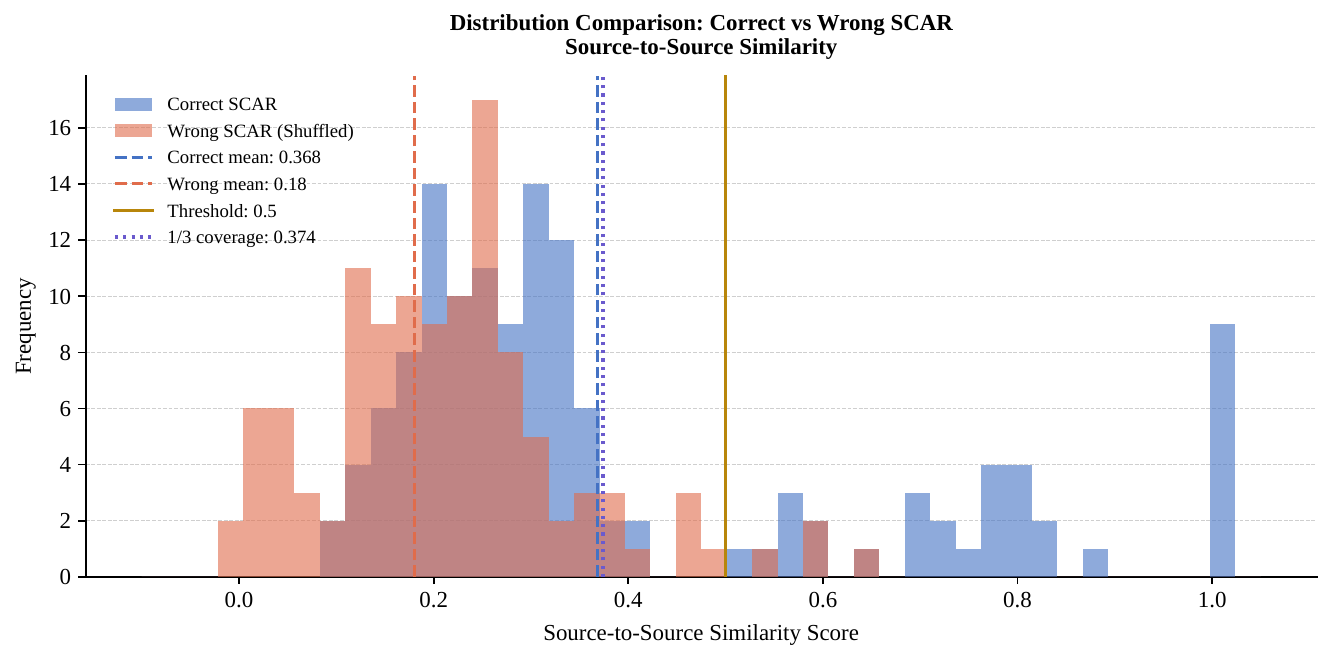}
    \caption{Distribution of source-to-source similarity scores in SCAR for correct (gold) and incorrect (shuffled) pairs. Dashed lines indicate the mean similarity for each group, while the dotted line shows the chosen threshold (0.374), corresponding to the upper tertile of gold similarities. A higher fixed threshold (e.g., 0.5) would exclude many valid analogies, motivating the use of a data-driven cutoff.}
    \label{fig:threshold}
\end{figure}

Figure~\ref{fig:threshold} shows that correct source pairs tend to exhibit higher similarity than incorrect ones, but with substantial overlap between the two distributions. This overlap indicates that a fixed high threshold (e.g., 0.5) would be overly restrictive, excluding many valid analogies that are structurally sound but not highly similar at the surface level.

To balance precision and coverage, we adopt a data-driven threshold of 0.374, corresponding to the upper tertile (top one-third) of similarity scores among gold source pairs. This choice captures a substantial portion of valid analogies while filtering out a large fraction of incorrect ones. As shown in the figure, this threshold lies above the mean similarity of incorrect pairs (0.180) and close to the mean of correct pairs (0.368), providing a principled trade-off between false positives and false negatives.

\section{Explanation Generation Prompts}
\label{app:explanation-prompt}

This appendix lists the \texttt{DSPy} signatures used for the six explanation-generation settings. Each signature corresponds to one input configuration.

\paragraph{Setting 1: System Names Only}
This configuration provides only the target and source system names.

\begin{lstlisting}[language=Python]
class ExplGen_Base(dspy.Signature):
    """Generate an analogy explanation using system names only."""
    target: str = dspy.InputField(
        desc="Target system"
    )
    source: str = dspy.InputField(
        desc="Source system"
    )
    explanation: str = dspy.OutputField(
        desc="Explanation of the analogy"
    )
\end{lstlisting}

\paragraph{Setting 2: System Names with Background Descriptions}
This setting augments system names with background descriptions for both systems.

\begin{lstlisting}[language=Python]
class ExplGen_BaseDesc(dspy.Signature):
    """Generate an analogy explanation using names and descriptions."""
    target: str = dspy.InputField(
        desc="Target system"
    )
    target_desc: str = dspy.InputField(
        desc="Background description of the target"
    )
    source: str = dspy.InputField(
        desc="Source system"
    )
    source_desc: str = dspy.InputField(
        desc="Background description of the source"
    )
    explanation: str = dspy.OutputField(
        desc="Explanation of the analogy"
    )
\end{lstlisting}

\paragraph{Setting 3: System Names with Unpaired Sub-concepts}
This configuration provides both systems with unpaired lists of sub-concepts.

\begin{lstlisting}[language=Python]
class ExplGen_Sub(dspy.Signature):
    """Generate an analogy explanation using unpaired sub-concepts."""
    target: str = dspy.InputField(
        desc="Target system"
    )
    target_sub: list[str] = dspy.InputField(
        desc="Target sub-concepts"
    )
    source: str = dspy.InputField(
        desc="Source system"
    )
    source_sub: list[str] = dspy.InputField(
        desc="Source sub-concepts"
    )
    explanation: list[str] = dspy.OutputField(
        desc="Explanation after implicitly pairing sub-concepts"
    )
\end{lstlisting}

\paragraph{Setting 4: System Names with Unpaired Sub-concepts and Descriptions}
This setting extends Setting~3 by adding background descriptions for both systems.

\begin{lstlisting}[language=Python]
class ExplGen_SubDesc(dspy.Signature):
    """Generate an analogy explanation using unpaired sub-concepts and descriptions."""
    target: str = dspy.InputField(
        desc="Target system"
    )
    target_desc: str = dspy.InputField(
        desc="Background description of the target"
    )
    target_sub: list[str] = dspy.InputField(
        desc="Target sub-concepts"
    )
    source: str = dspy.InputField(
        desc="Source system"
    )
    source_desc: str = dspy.InputField(
        desc="Background description of the source"
    )
    source_sub: list[str] = dspy.InputField(
        desc="Source sub-concepts"
    )
    explanation: list[str] = dspy.OutputField(
        desc="Explanation after implicitly pairing sub-concepts"
    )
\end{lstlisting}

\paragraph{Setting 5: Paired Sub-concept Mappings}
This configuration provides explicit one-to-one sub-concept alignments between the target and source systems.

\begin{lstlisting}[language=Python]
class ExplGen_Pair(dspy.Signature):
    """Generate an analogy explanation using paired sub-concepts."""
    target: str = dspy.InputField(
        desc="Target system"
    )
    source: str = dspy.InputField(
        desc="Source system"
    )
    pairs: list[list[str]] = dspy.InputField(
        desc="Paired target-source sub-concepts"
    )
    explanation: list[str] = dspy.OutputField(
        desc="Explanation for each paired sub-concept"
    )
\end{lstlisting}

\paragraph{Setting 6: Paired Sub-concept Mappings with Descriptions}
This setting combines explicit sub-concept pairings with background descriptions for both systems.

\begin{lstlisting}[language=Python]
class ExplGen_PairDesc(dspy.Signature):
    """Generate an analogy explanation using paired sub-concepts and descriptions."""
    target: str = dspy.InputField(
        desc="Target system"
    )
    target_desc: str = dspy.InputField(
        desc="Background description of the target"
    )
    source: str = dspy.InputField(
        desc="Source system"
    )
    source_desc: str = dspy.InputField(
        desc="Background description of the source"
    )
    pairs: list[list[str]] = dspy.InputField(
        desc="Paired target-source sub-concepts"
    )
    explanation: list[str] = dspy.OutputField(
        desc="Explanation for each paired sub-concept"
    )
\end{lstlisting}

\section{Prompt for the LLM-as-a-Judge}
\label{app:llm-as-judge}
\begin{lstlisting}[caption={Prompt used to evaluate the analogies from different settings on different dimensions. This was wrapped in DSPY function as instruction, alongside specific input and output variables.}]
You are an expert evaluator of scientific analogies.

Given a target concept and a chosen source analogy, evaluate whether this is a good analogy.
A good analogy uses a FAMILIAR source concept to explain an UNFAMILIAR target concept through
meaningful structural or functional parallels.

For each of the three dimensions below, first provide a brief reasoning explaining your
assessment, then give the numeric score (1, 2, or 3).

ANALOGY_COHERENCE: Does the pairing make intuitive sense?
- 3: The connection is immediately clear and natural -- most people would see it without explanation.
     The source and target share an obvious structural or functional parallel.
- 2: A meaningful connection exists but requires some explanation to see.
     The link is real but not self-evident; a sentence or two is needed to establish it.
- 1: No meaningful connection exists, or the pairing is random, forced, or misleading.

MAPPING_SOUNDNESS: Could properties/mechanisms of the source map to the target?
- 3: Rich, consistent structural or functional correspondences exist.
     Multiple source properties map precisely onto target properties.
     The mapping holds across the main components of both domains.
- 2: Some valid mappings exist, but coverage is partial.
     Core correspondences work, but important aspects of the target are not represented,
     or some mappings are approximate or strained.
- 1: No valid mappings are possible, or the apparent mappings are entirely superficial
     or misleading. Source and target are fundamentally incompatible.

EXPLANATORY_POWER: Would this analogy help a learner understand the target?
- 3: The analogy clearly illuminates the target concept and supports correct reasoning.
     A learner could use it to predict or explain target behavior.
- 2: The analogy provides partial insight with notable limitations.
     It conveys the general idea but cannot support deeper reasoning,
     or it risks creating minor misconceptions.
- 1: The analogy fails to aid understanding and would likely confuse or mislead a learner.

CALIBRATION EXAMPLES -- use these to anchor your scoring.
Score each dimension independently. The scale is 1-3 only:
  3 = strong / clearly works
  2 = partial / works with caveats
  1 = doesn't work / poor or misleading

Example 1 (scores: 3 / 3 / 3)
  target: "electric circuit"  |  analogy: "water flowing through pipes"

  coherence=3:
    The connection is immediately clear without any explanation needed.
    A driving force pushes a substance through a constrained pathway in both cases --
    this structural parallel is instantly visible.

  mapping=3:
    Multiple precise correspondences hold consistently:
      voltage -> pressure | current -> flow rate | resistance -> pipe restriction
      battery -> pump | closed loop -> closed pipe system
    Relationships between variables (increase pressure -> increase flow) are preserved.

  explanatory=3:
    A learner can use the analogy to reason about Ohm's law qualitatively
    and make correct predictions about circuit behavior.

Example 2 (scores: 3 / 2 / 3)
  target: "cell"  |  analogy: "factory"

  coherence=3:
    The analogy captures organized division of labor within a bounded system.
    Most people immediately see why a cell resembles a factory.

  mapping=2:
    nucleus -> control center | ribosomes -> production workers
    mitochondria -> energy supply | membrane -> boundary/security

  explanatory=3:
    Helps learners grasp coordination and specialization clearly.

Example 3 (scores: 2 / 2 / 2)
  target: "mathematical function"  |  analogy: "machine"

  coherence=2:
    The shared idea (input -> transformation -> output) is real but requires explanation.

  mapping=2:
    input -> raw material, rule -> mechanism, output -> product

  explanatory=2:
    Useful for initial intuition but limited for deeper reasoning.

Example 4 (scores: 1 / 1 / 1)
  target: "neural network"  |  analogy: "human brain"

  coherence=1:
    Superficial similarity ("neurons") without structural alignment.

  mapping=1:
    No valid correspondences across core mechanisms.

  explanatory=1:
    Likely to mislead understanding.

Example 5 (scores: 1 / 1 / 1)
  target: "chemical reaction"  |  analogy: "a novel"

  coherence=1:
    No meaningful connection.

  mapping=1:
    No systematic correspondences.

  explanatory=1:
    No instructional value.

Example 6 (scores: 3 / 1 / 2)
  target: "democracy"  |  analogy: "majority vote in a classroom"

  coherence=3:
    Both involve majority-based decisions.

  mapping=1:
    Institutional aspects have no counterpart.

  explanatory=2:
    Partial understanding only.

Example 7 (scores: 2 / 3 / 3)
  target: "atom"  |  analogy: "solar system"

  coherence=2:
    Spatial similarity exists but requires explanation.

  mapping=3:
    nucleus -> sun | electrons -> planets

  explanatory=3:
    Supports initial reasoning about structure.

Example 8 (scores: 3 / 2 / 1)
  target: "photosynthesis"  |  analogy: "solar-powered factory"

  coherence=3:
    Energy conversion is clearly analogous.

  mapping=2:
    sunlight -> energy supply | chloroplast -> factory

  explanatory=1:
    Oversimplifies processes.

Example 9 (scores: 1 / 3 / 3)
  target: "compound interest"  |  analogy: "a snowball rolling downhill"

  coherence=1:
    Domains differ fundamentally.

  mapping=3:
    principal -> snowball | growth -> accumulation

  explanatory=3:
    Clearly conveys accelerating growth.

Example 10 (scores: 3 / 1 / 2)
  target: "ecosystem"  |  analogy: "a family"

  coherence=3:
    Interdependence is clear.

  mapping=1:
    No structural ecological correspondences.

  explanatory=2:
    Partial intuition only.
\end{lstlisting}

\section{Ablation results for Sub-concept Matching}
\label{app:ablation}

\begin{figure}[h]
    \centering
    \includegraphics[width=\linewidth]{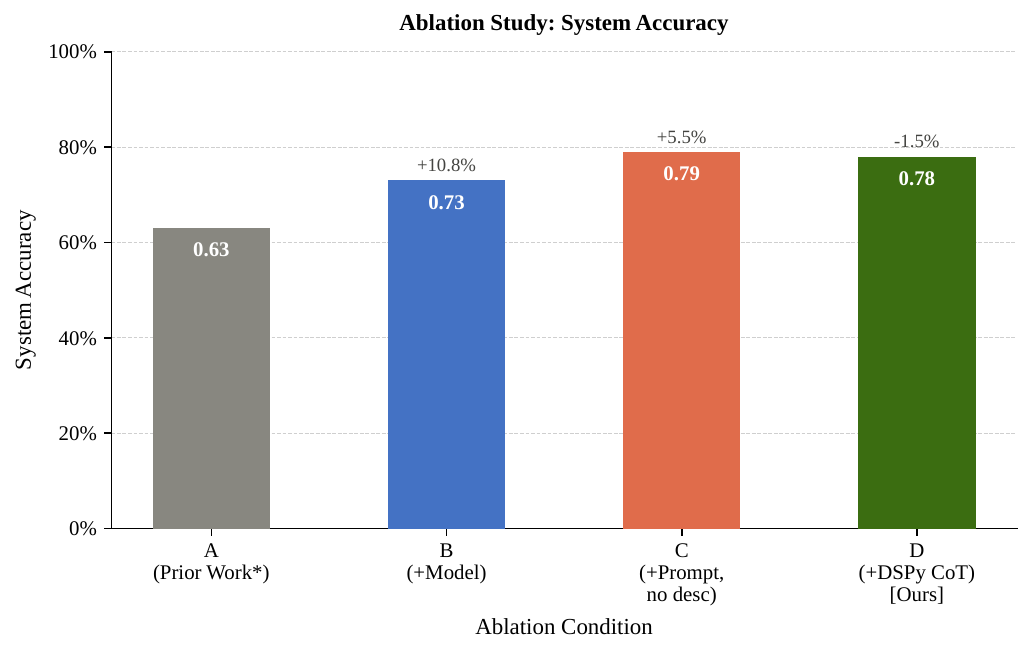}
    \caption{Ablation study on sub-concept matching system accuracy, isolating the contribution of each component relative to prior work~\cite{yuan2023beneath}. Upgrading the model to our best preforming Grok-4 alone (B) yields the largest gain (+10.8\%), followed by prompt redesign (C, +5.5\%). Wrapping with DSPy Chain-of-Thought (D, our final system) introduces a marginal drop ($-$1.5\%), confirming that automated prompt optimisation does not improve over the hand-crafted prompt for this task.}
    \label{fig:abalation}
\end{figure}

Figure \ref{fig:abalation} shows the ablation study to figure out what aspect actually led to the improved results.

\section{Explanation Heatmap Results}
\label{app:explanation_heat}

\begin{figure}[h]
    \centering
    \includegraphics[width=\linewidth]{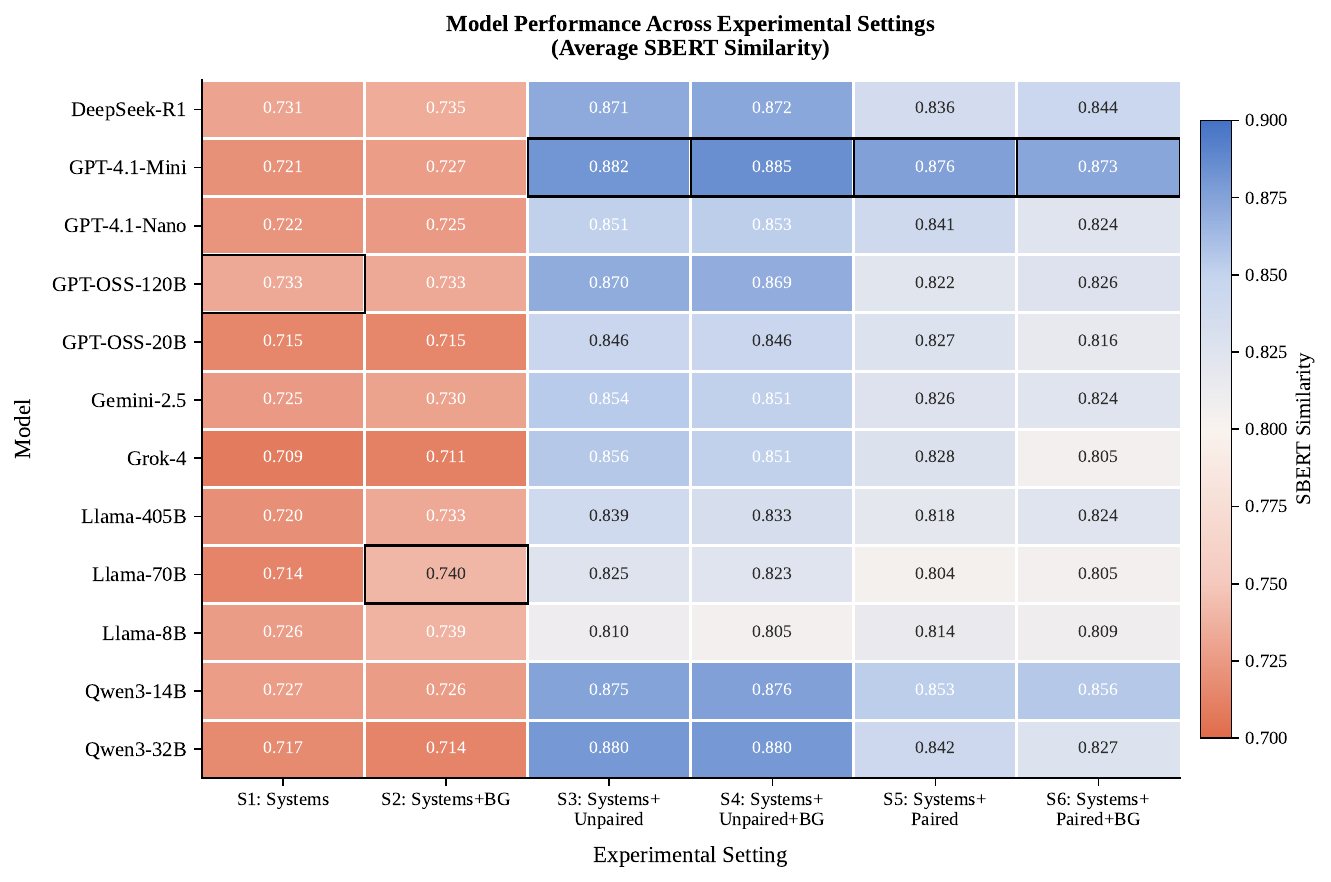}
    \caption{Average SBERT similarity scores across twelve models and six experimental settings. Settings S1-S2 provide only system-level context (without and with background descriptions, respectively), while S3-S4 additionally supply unpaired sub-concepts and S5-S6 supply paired sub-concepts. Black borders indicate the best-performing model per setting. GPT-4.1-Mini consistently leads in sub-concept-enriched settings (S3-S5), while DeepSeek-R1 and Qwen3 variants are competitive in paired settings (S5-S6). Scores rise substantially when sub-concepts are introduced (S3-S6 vs.\ S1-S2), suggesting that grounding the comparison in explicit sub-concept structure is a stronger driver of similarity than background context alone.}
    \label{fig:explanation_heatmap}
\end{figure}

Figure~\ref{fig:explanation_heatmap} presents SBERT similarity scores across twelve LLMs and six experimental settings for the explanation generation stage.

\section{Open Setting Source Finding: Per-Model Hit@20 Comparison }
\label{app:targetvssubconcept}
\begin{figure}[h]
    \centering
    \includegraphics[width=\linewidth]{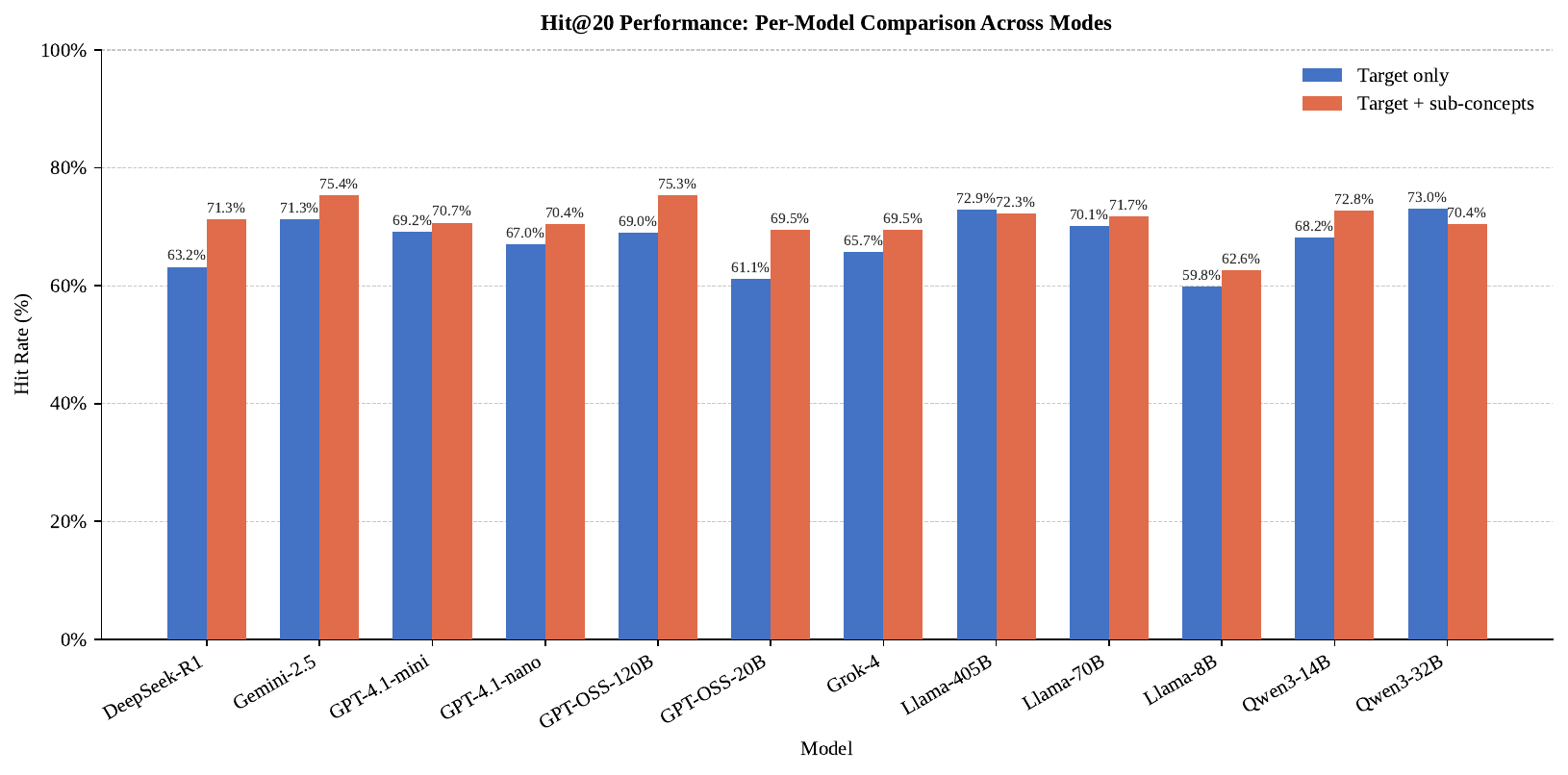}
    \caption{Per-model Hit@20 comparison in the open setting under Target-Only and Target + Sub-Concept modes. Results are reported using semantic matching and highlight the effect of incorporating sub-concepts on top-20 retrieval performance.}
    \label{fig:hit_20_per_model_comparison}
\end{figure}

Figure~\ref{fig:hit_20_per_model_comparison} presents per-model Hit@20 scores in the open setting, comparing target-only and target with sub-concept prompting across all twelve models.

\section{Embedding-based Retrieval Results}
\label{app:retrieval_results}
\begin{figure}[h]
    \centering
    \includegraphics[width=\linewidth]{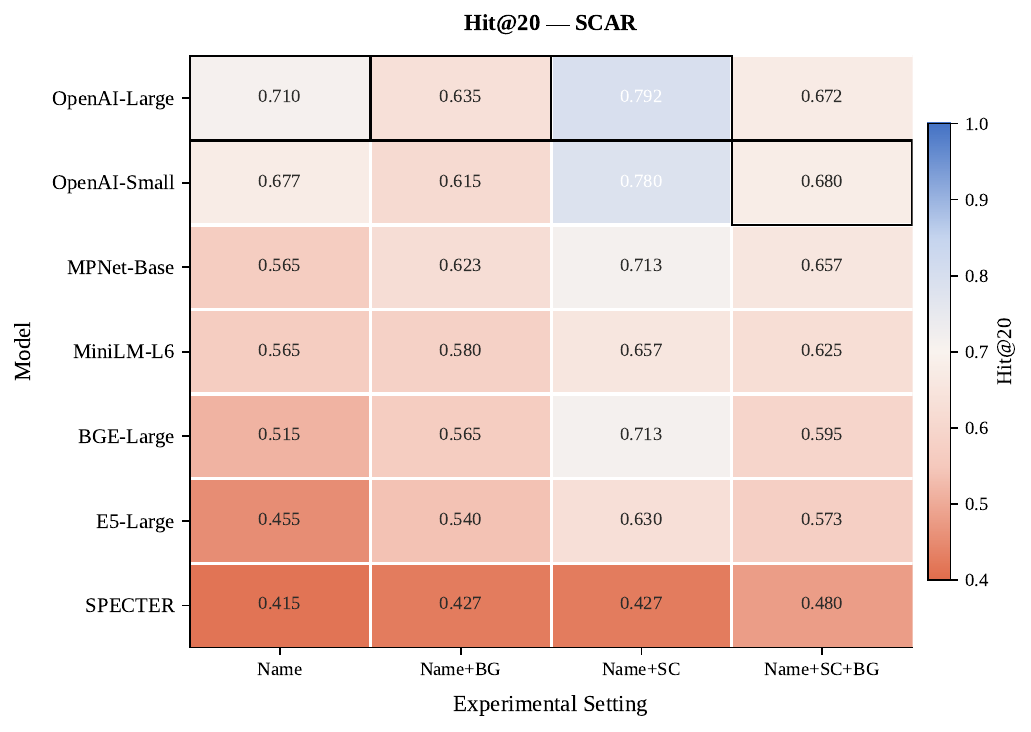}
    \caption{Hit@20 performance on the SCAR dataset across embedding models and input configurations. Incorporating sub-concepts consistently improves retrieval compared to name-only inputs, while adding background provides mixed benefits. OpenAI embeddings achieve the strongest overall performance.}
    \label{fig:hit20_scar}
\end{figure}

\begin{figure}[h]
    \centering
    \includegraphics[width=\linewidth]{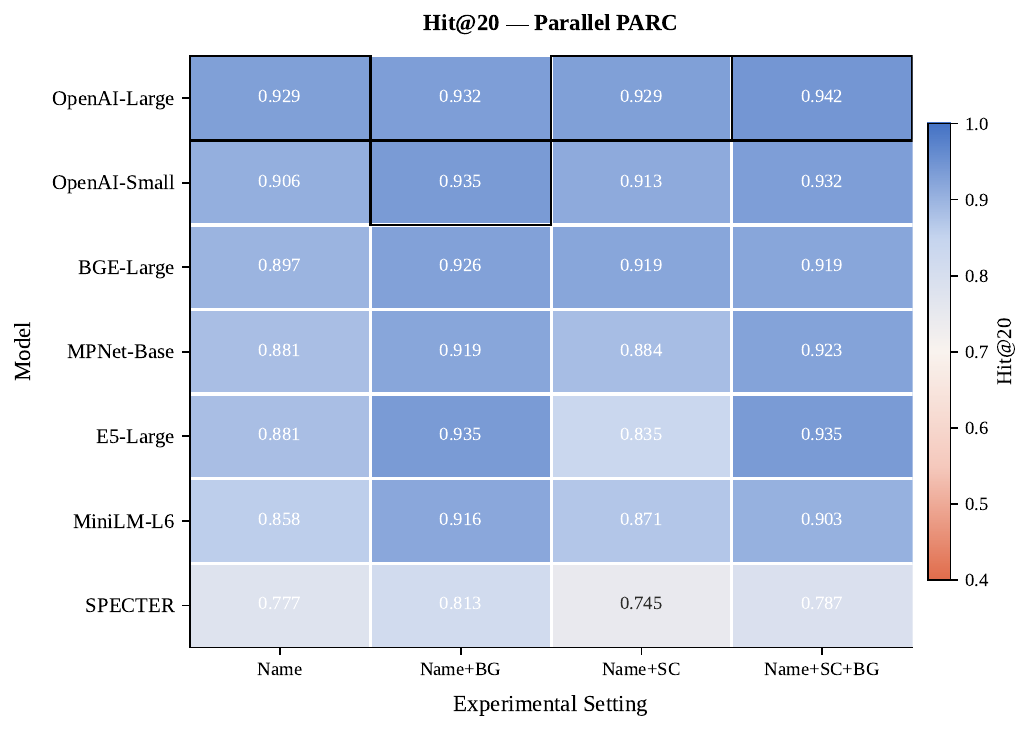}
    \caption{Hit@20 performance on the ParallelPARC dataset. Retrieval performance is generally higher than SCAR, reflecting the more structured nature of process-based analogies. Combining sub-concepts with background yields the best results across most models, indicating the value of richer contextual representations.}
    \label{fig:hit20_parc}
\end{figure}

Figure \ref{fig:hit20_scar}  and \ref{fig:hit20_parc} show the heatmap of the 7 embedding models across 4 input configurations.

\section{Annotation Dataset}
\label{app:45table}
\begin{figure}[p]
\centering
\includegraphics[width=\columnwidth]{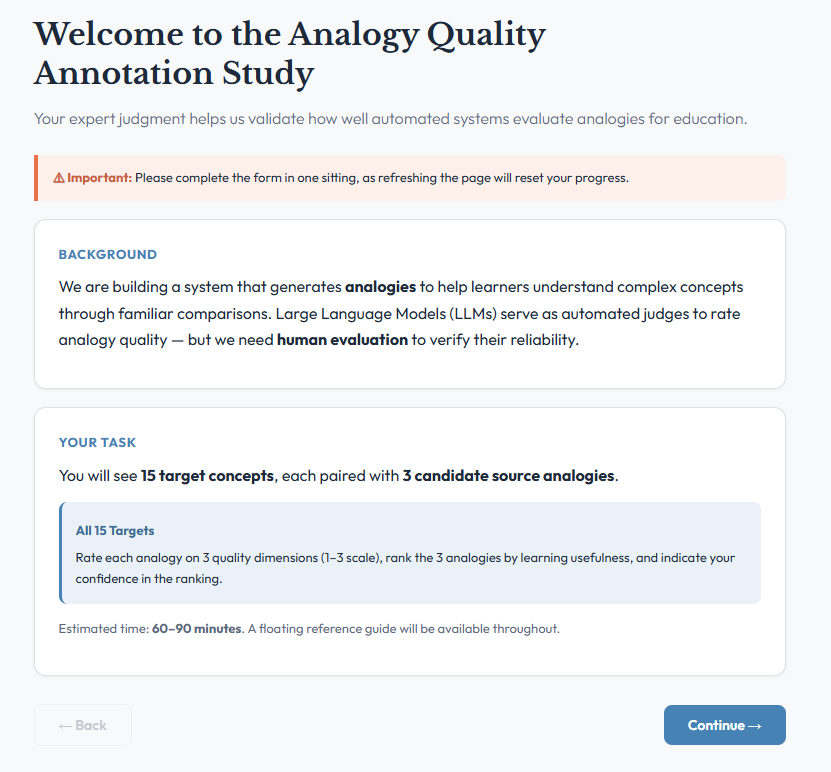}
\caption{(a) Welcome screen introducing the task and study background.}
\label{fig:annotation-form1}
\end{figure}

\begin{figure}[p]
\centering
\includegraphics[width=\columnwidth]{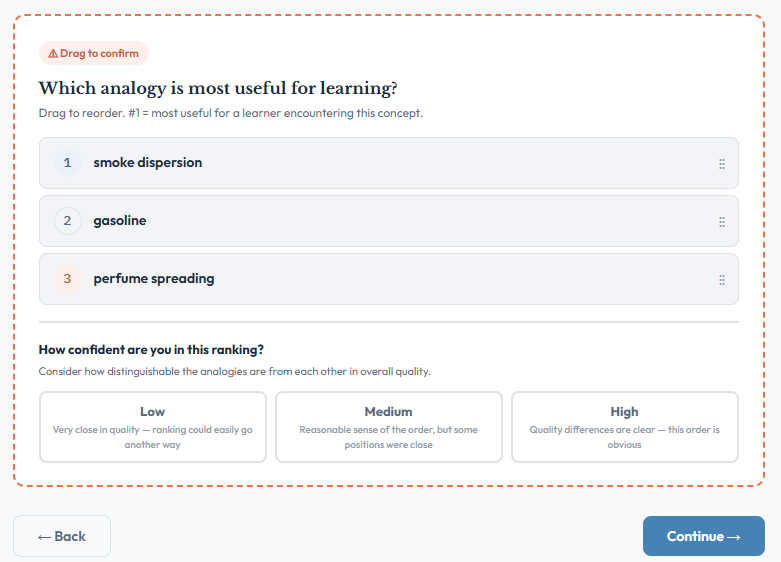}
\caption{(j) Drag-to-rank interface for ordering sources by learning usefulness with confidence selection.}
\label{fig:annotation-form}
\end{figure}

Table~\ref{tab:annotation-data} presents the annotation dataset, comprising 15 target concepts spanning five domains, each paired with three source analogies evaluated under target-only and target with sub-concept conditions. Furthermore, Figures \ref{fig:annotation-form1}-~\ref{fig:annotation-form} provide an overview of the web-based annotation form used in our human evaluation study. The form guides annotators through rating each of three candidate source analogies per target concept on three dimensions -- coherence, mapping soundness, and explanatory power -- on a 1--3 scale, before ranking the three sources by overall learning usefulness and reporting their confidence in the ranking.

\begin{table}[t]
\centering
\caption{Annotation dataset: 15 target concepts across five domains, each paired with three source analogies. T = Target-only, T+S = Target + Sub-concepts.}
\label{tab:annotation-data}
\small
\setlength{\tabcolsep}{4pt}
\begin{tabular}{|c|l|l|l|l|c|}
\hline
\textbf{\#} & \textbf{Target} & \textbf{Domain} & \textbf{Source Analogy} & \textbf{Model} & \textbf{Mode} \\
\hline
1 & Gas Diffusion & Chemistry & Perfume spreading & GPT-OSS-120B & T \\
 & & & Smoke dispersion & Qwen3-14B & T+S \\
 & & & Gasoline & Qwen3-14B & T \\
\hline
2 & Encryption & CS & locked box & GPT-OSS-120B & T \\
 & & & Sealed envelope & Grok-4-fast & T+S \\
 & & & Locker & GPT-4.1-nano & T+S \\
\hline
3 & Industrial Revolution & History & Steam engine era & GPT-OSS-120B & T \\
 & & & Automation & LLaMA-8B & T+S \\
 & & & Industrialization & LLaMA-405B & T \\
\hline
4 & GAN Algorithm & CS & Art forgery & Grok-4-fast & T \\
 & & & Artist and Critic & Grok-4-fast & T+S \\
 & & & Teacher & GPT-4.1-nano & T \\
\hline
5 & Bill of Rights & Law & Rulebook & Qwen3-32B & T \\
 & & & New‑car warranty & GPT-OSS-20B & T \\
 & & & School curfew & GPT-OSS-20B & T \\
\hline
6 & AdaBoost Algorithm & CS & Teamwork & Gemini-2.5 & T+S \\
 & & & iterative training & GPT-OSS-120B & T \\
 & & & Gears in a machine & LLaMA-70B & T+S \\
\hline
7 & Platelets & Biology & Tiny repair crew & Gemini-2.5 & T \\
 & & & Band-Aid & Grok-4-fast & T+S \\
 & & & Beads & GPT-4.1-mini & T \\
\hline
8 & Cell & Biology & factory & Grok-4-fast & T \\
 & & & House & LLaMA-405B & T \\
 & & & Garage & LLaMA-8B & T+S \\
\hline
9 & Economics & Economics & Marketplace & Qwen3-14B & T \\
 & & & Household Budget & LLaMA-70B & T \\
 & & & Ecosystem & Qwen3-14B & T+S \\
\hline
10 & Chemical Reaction & Chemistry & Baking soda + vinegar & Qwen3-14B & T \\
 & & & Digestion & Qwen3-32B & T+S \\
 & & & Boiling water & GPT-4.1-mini & T \\
\hline
11 & Photosynthesis & Biology & Solar panels & LLaMA-405B & T+S \\
 & & & Greenhouse farming & Grok-4-fast & T \\
 & & & Fermentation Process & LLaMA-70B & T+S \\
\hline
12 & Random forest & CS & Voting committee & Gemini-2.5 & T+S \\
 & & & Jury deliberation & GPT-OSS-120B & T+S \\
 & & & Tree & GPT-4.1-nano & T \\
\hline
13 & Isotope Dating & Chemistry & Hourglass & DeepSeek-R1 & T+S \\
 & & & Atomic Clock & LLaMA-405B & T \\
 & & & Half-Life & LLaMA-70B & T \\
\hline
14 & Renaissance & History & spring bloom & GPT-OSS-120B & T \\
 & & & Ancient Rome & retrieval-based & Embedding \\
 & & & Enlightenment & LLaMA-70B & T+S \\
\hline
15 & EM Algorithm & CS & Detective investigation & GPT-OSS-120B & T \\
 & & & guess‑and‑check & GPT-OSS-120B & T+S \\
 & & & Miner & retrieval-based & Embedding \\
\hline
\end{tabular}
\end{table}

\clearpage

\begin{figure}[p]
\centering
\includegraphics[width=\columnwidth]{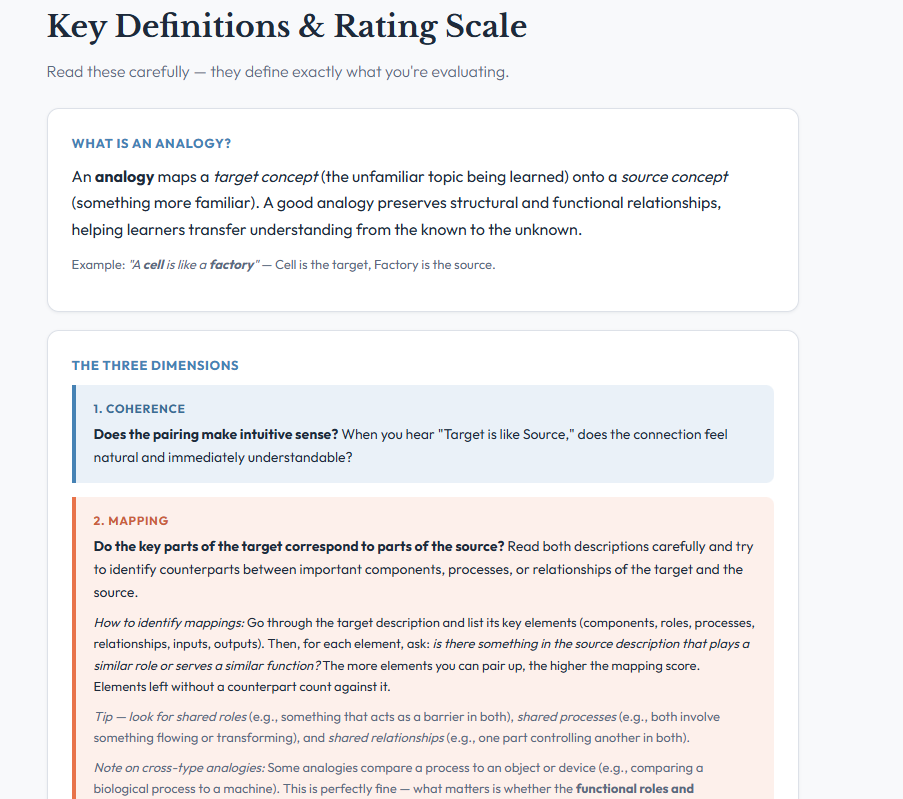}
\caption{(b) Key definitions and rating scale for the three evaluation dimensions.}
\end{figure}
\clearpage

\begin{figure}[p]
\centering
\includegraphics[width=\columnwidth]{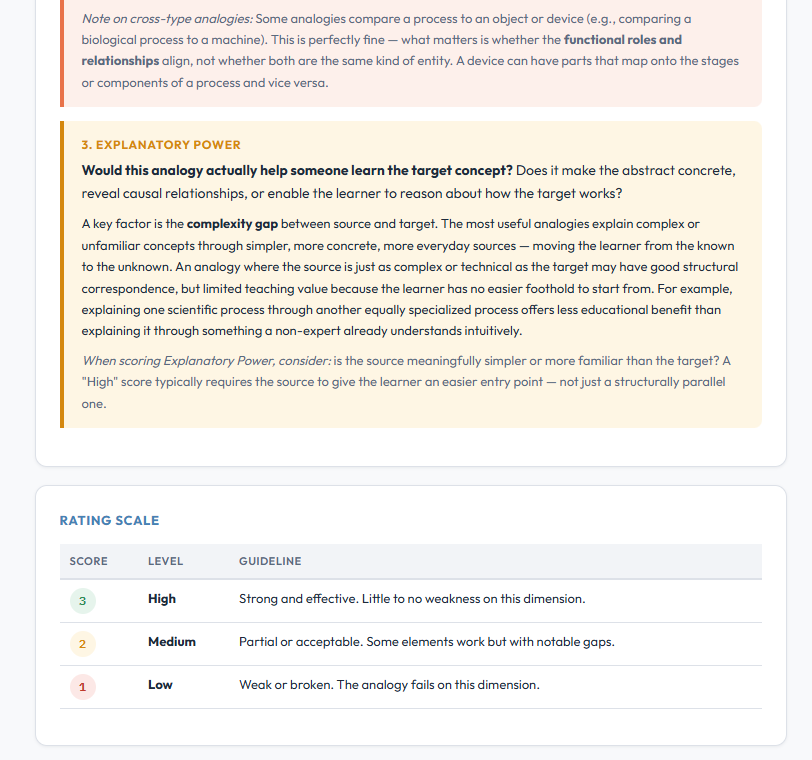}
\caption{(c) Explanatory power dimension definition and full rating scale.}
\end{figure}
\clearpage

\begin{figure}[p]
\centering
\includegraphics[width=\columnwidth]{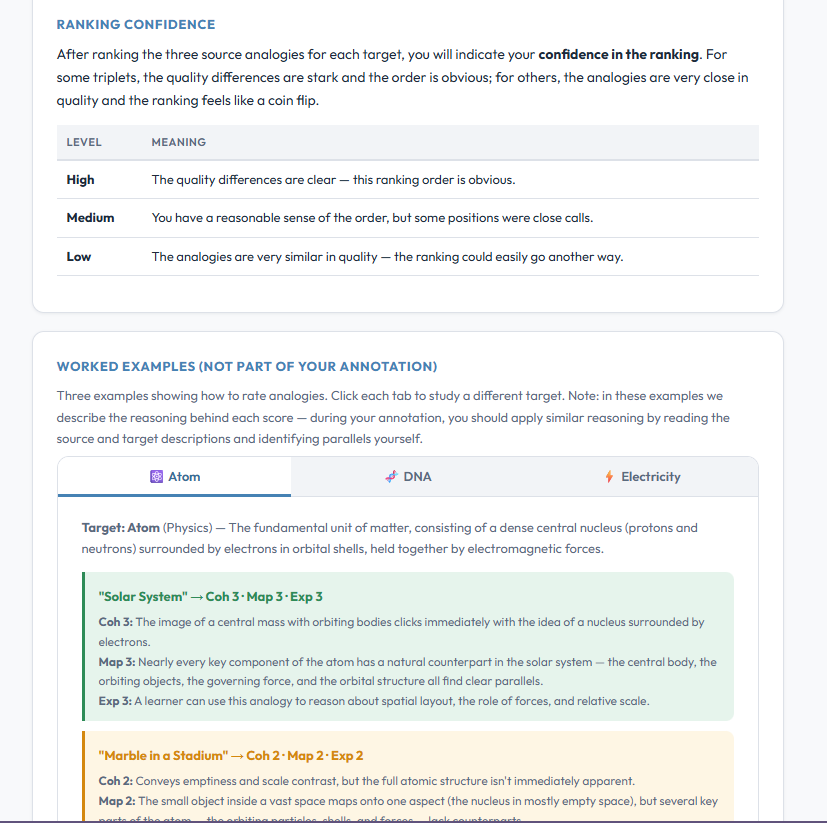}
\caption{(d) Ranking confidence scale and worked examples panel.}
\end{figure}
\clearpage

\begin{figure}[p]
\centering
\includegraphics[width=\columnwidth]{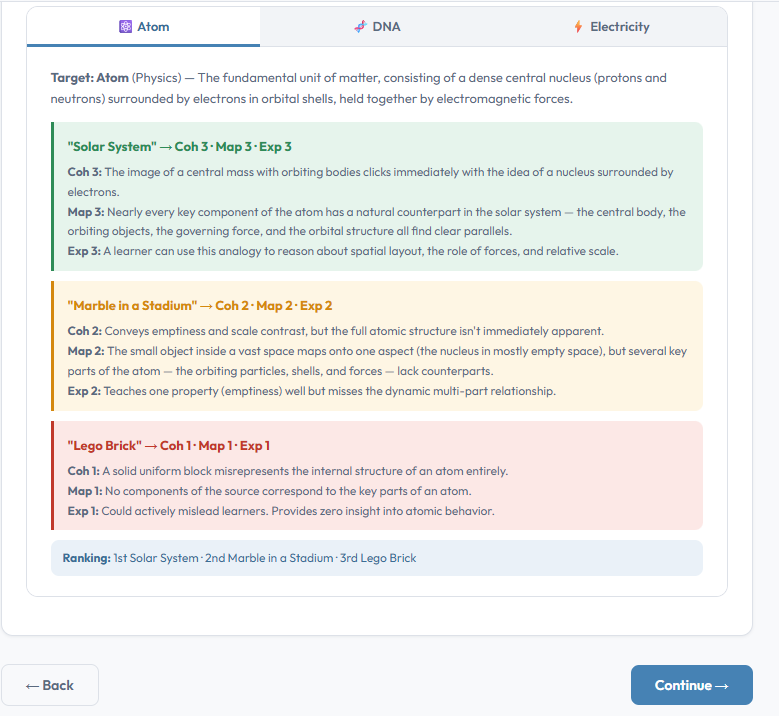}
\caption{(e) Worked example for the \textit{Atom} target with scored candidate sources.}
\end{figure}
\clearpage

\begin{figure}[p]
\centering
\includegraphics[width=\columnwidth]{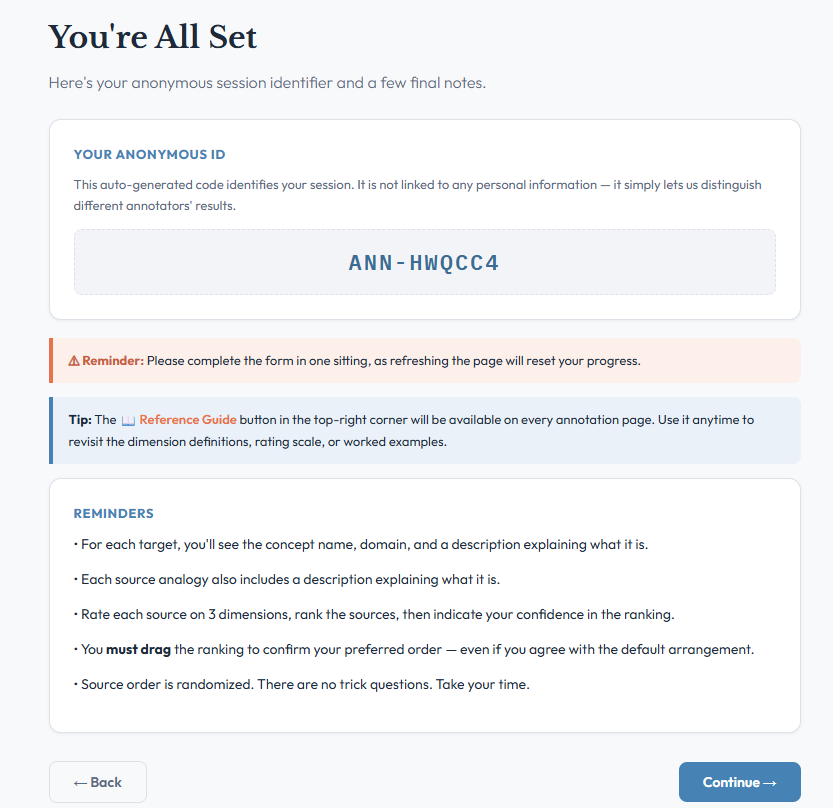}
\caption{(f) Annotator onboarding screen with anonymous session ID and task reminders.}
\end{figure}
\clearpage

\begin{figure}[p]
\centering
\includegraphics[width=\columnwidth]{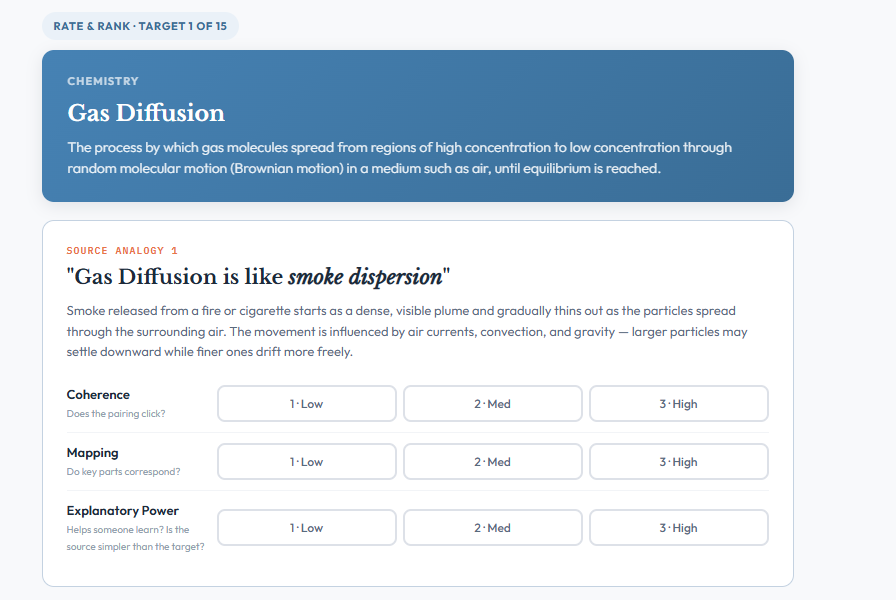}
\caption{(g) Rating interface for \textit{Gas Diffusion} — Source Analogy 1.}
\end{figure}
\clearpage

\begin{figure}[p]
\centering
\includegraphics[width=\columnwidth]{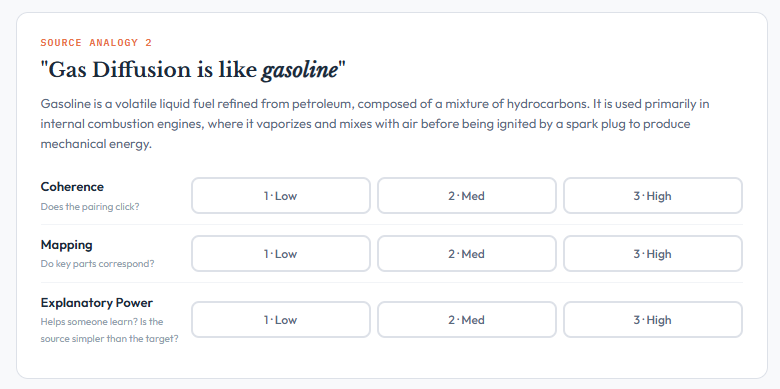}
\caption{(h) Rating interface for \textit{Gas Diffusion} — Source Analogy 2.}
\end{figure}
\clearpage

\begin{figure}[p]
\centering
\includegraphics[width=\columnwidth]{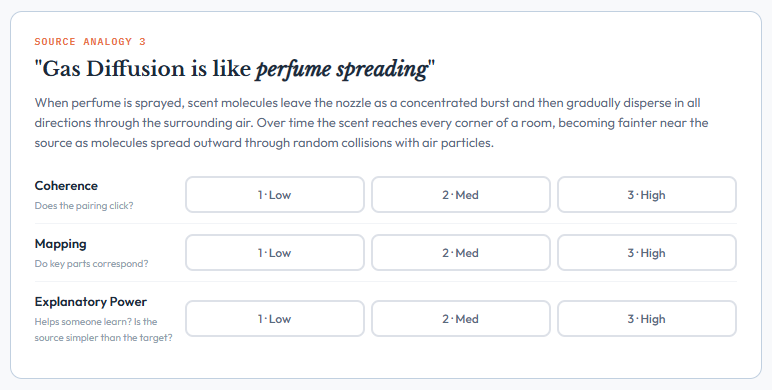}
\caption{(i) Rating interface for \textit{Gas Diffusion} — Source Analogy 3.}
\end{figure}
\clearpage

\clearpage

\section{Human Evaluation Rank Agreement}
\label{app:human_agreement}

\begin{table}[h]
\centering
\small
\resizebox{\textwidth}{!}{%
\begin{tabular}{lccc|cc}
\hline
\textbf{Target} & \textbf{Analogy 1} & \textbf{Analogy 2} & \textbf{Analogy 3} & \textbf{Kendall’s W} & \textbf{p-value} \\
\hline
adaboost algorithm & gears in a machine & iterative training & teamwork & 0.735 & 0.0058 \\
bill of rights & new-car warranty & rulebook & school curfew & 0.388 & 0.0663 \\
cell & factory & garage & house & 0.551 & 0.0211 \\
chemical reaction & baking soda + vinegar & boiling water & digestion & 0.878 & 0.0021 \\
economics & ecosystem & household budget & marketplace & 0.020 & 0.8669 \\
em algorithm & detective investigation & guess-and-check & miner & 0.755 & 0.0051 \\
encryption & locked box & locker & sealed envelope & 0.143 & 0.3679 \\
gan algorithm & art forgery & artist and critic & teacher & 0.878 & 0.0021 \\
gas diffusion & gasoline & perfume spreading & smoke dispersion & 0.796 & 0.0038 \\
industrial revolution & automation & industrialization & steam engine era & 0.061 & 0.6514 \\
isotope dating & atomic clock & half-life & hourglass & 0.633 & 0.0119 \\
photosynthesis & fermentation process & greenhouse farming & solar panels & 0.388 & 0.0663 \\
platelets & band-aid & beads & tiny repair crew & 1.000 & 0.0009 \\
random forest & jury deliberation & tree & voting committee & 1.000 & 0.0009 \\
renaissance & ancient rome & enlightenment & spring bloom & 0.061 & 0.6514 \\
\hline
\end{tabular}
}
\caption{Inter-annotator agreement for analogy ranking using Kendall’s W. Each row corresponds to a target concept with three candidate analogies. Higher $W$ indicates stronger agreement among annotators, while the p-value tests statistical significance.}
\label{tab:kendall_w}
\end{table}

Table~\ref{tab:kendall_w} reports inter-annotator agreement using Kendall’s coefficient of concordance (W) for each target concept. Annotators ranked three candidate analogies per target. Higher values of $W$ indicate stronger agreement.

\section{Full Human and LLM-as-a-judge Results Heatmap}
\label{full_human_judge}

\begin{figure}[h]
    \centering
    \includegraphics[width=\linewidth]{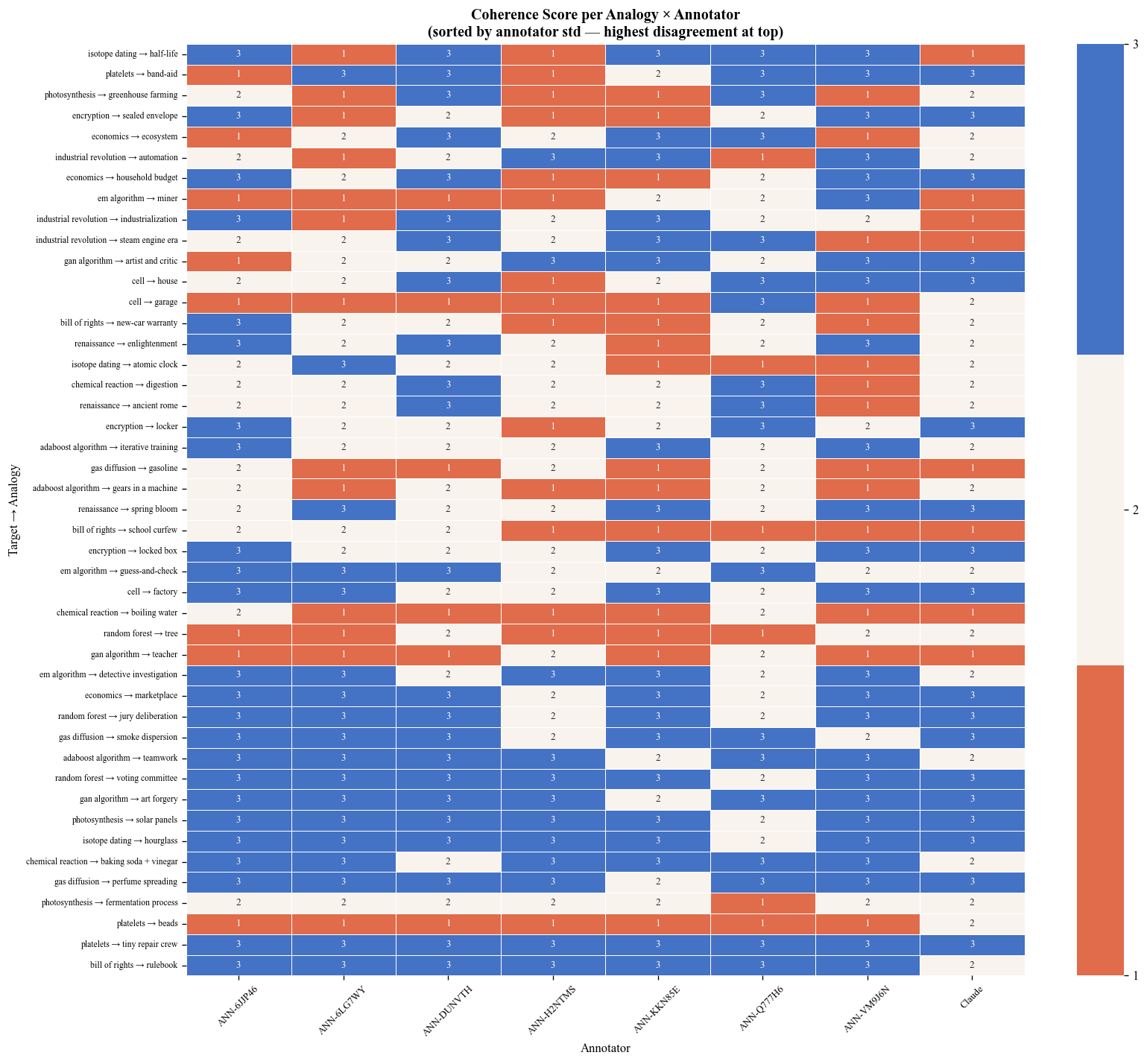}
    \caption{Mean coherence scores per analogy across eight annotators (seven human annotators and Claude), sorted by inter-annotator standard deviation (highest disagreement at top). Scores are on a 1--3 scale. High-disagreement analogies such as \textit{isotope dating -> half-life} and \textit{platelets -> band-aid} reveal cases where structural clarity is perceived differently across annotators. Low-disagreement analogies at the bottom (e.g., \textit{platelets -> tiny repair crew}, \textit{bill of rights -> rulebook}) receive consistently high scores, indicating clear and intuitive mappings. Claude's ratings (rightmost column) generally align with the human majority, with occasional deviations in mid-range cases.}
    \label{fig:coherence_all}
\end{figure}

\begin{figure}[h]
    \centering
    \includegraphics[width=\linewidth]{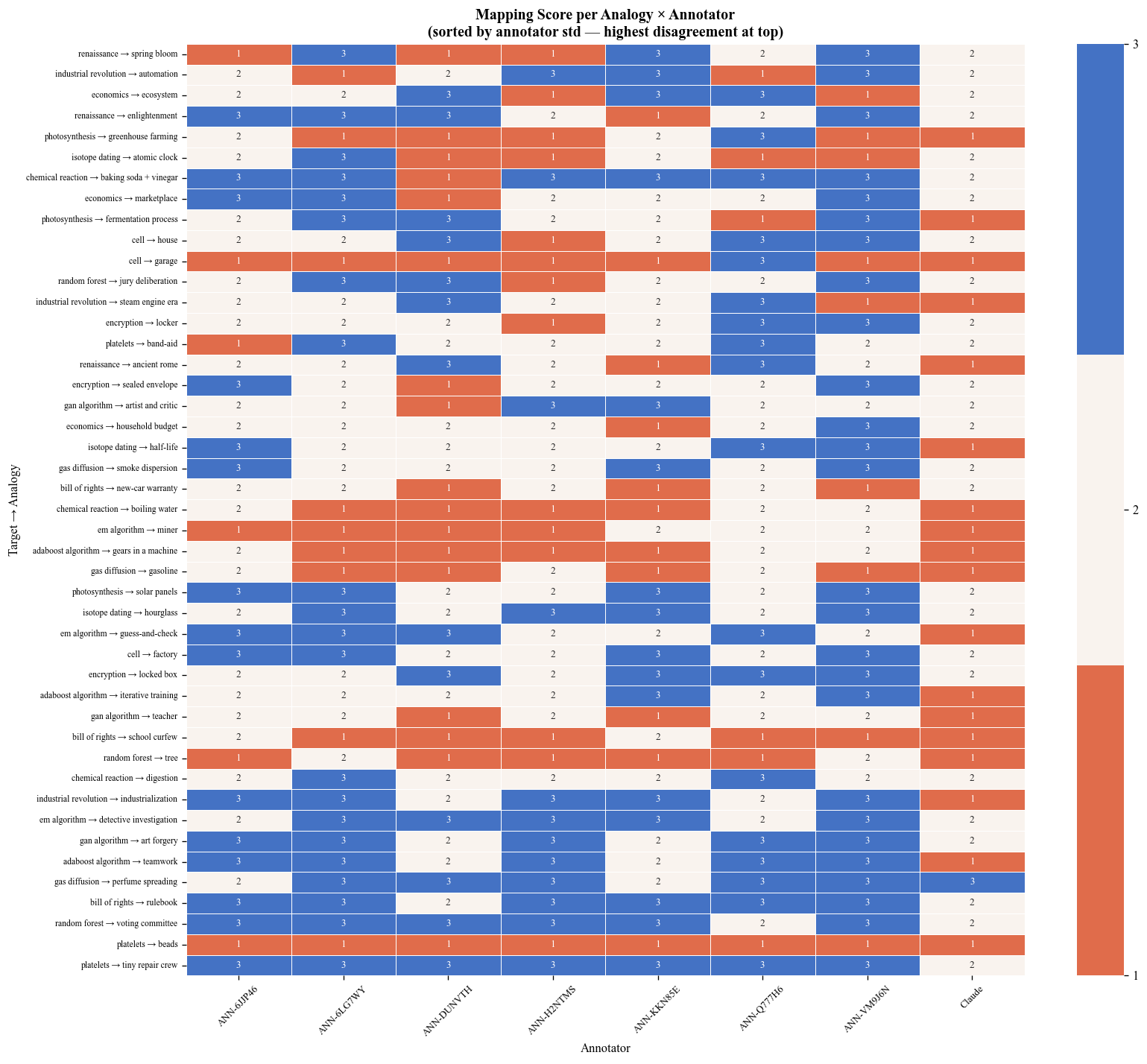}
    \caption{Mean mapping quality scores per analogy across eight annotators, sorted by inter-annotator standard deviation. Mapping quality—capturing how well source sub-concepts correspond to target sub-concepts—shows greater disagreement than coherence, particularly for analogies such as \textit{renaissance -> spring bloom} and \textit{industrial revolution -> automation}. Low-disagreement cases (e.g., \textit{platelets -> beads}, \textit{platelets -> tiny repair crew}) tend to receive uniformly low scores, suggesting that weak mappings are easier to agree on than strong ones. Claude's scores broadly follow human judgments, with a slight tendency toward mid-range values in ambiguous cases.}
    \label{fig:mapping_all}
\end{figure}

\begin{figure}[h]
    \centering
    \includegraphics[width=\linewidth]{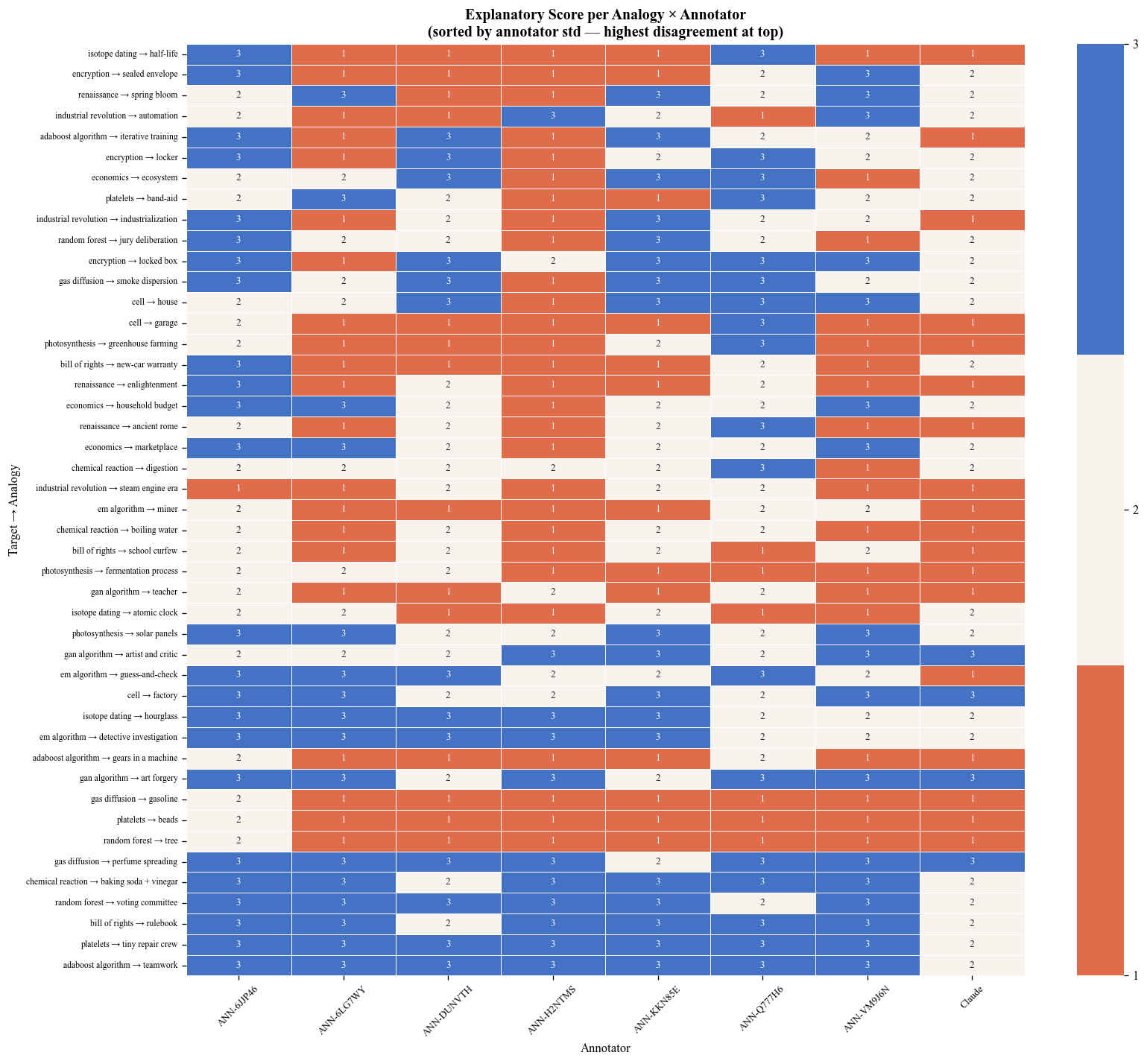}
    \caption{Mean explanatory power scores per analogy across eight annotators, sorted by inter-annotator standard deviation. Explanatory power—measuring how well an analogy supports understanding—exhibits the highest disagreement, especially for analogies such as \textit{isotope dating -> half-life} and \textit{encryption -> sealed envelope}. Consistently high-scoring analogies (e.g., \textit{platelets -> tiny repair crew}, \textit{adaboost algorithm -> teamwork}) indicate strong pedagogical value. Analogies involving less familiar concepts (e.g., \textit{gas diffusion -> gasoline}, \textit{random forest -> tree}) cluster in lower-score regions, suggesting more limited explanatory effectiveness.}
    \label{fig:explanatory_all}
\end{figure}

Figures \ref{fig:coherence_all}, \ref{fig:mapping_all} and \ref{fig:explanatory_all} show the full human and Claude sonnet 4.6 agreement.

\section{Word Overlap Analysis in Embedding-based Retrieval}
\begin{figure}[h]
    \centering
    \includegraphics[width=0.7\linewidth]{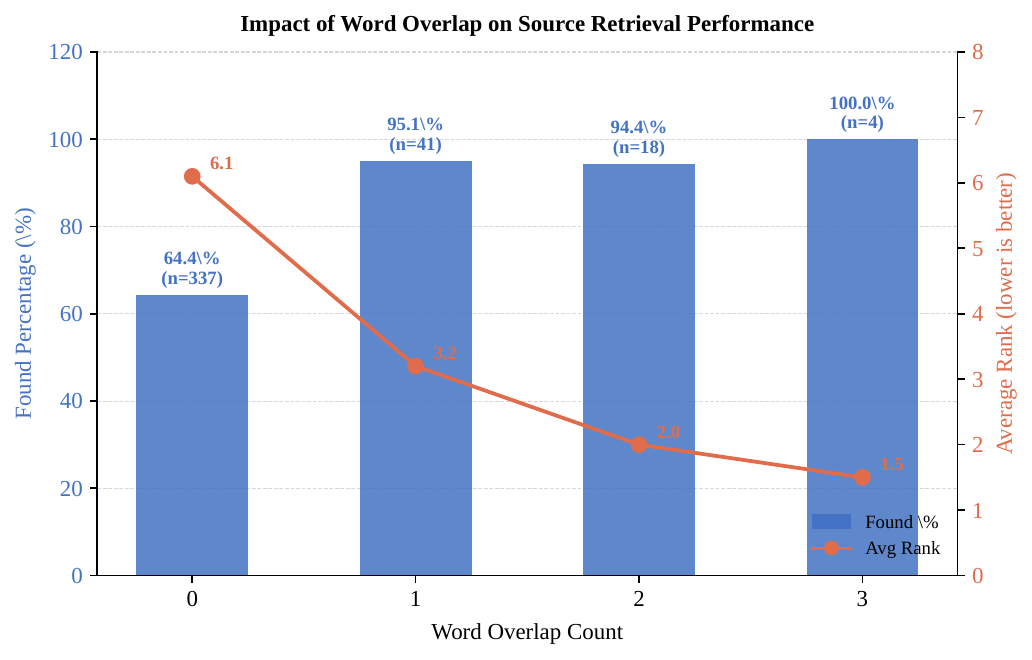}
    \caption{\textbf{Impact of lexical overlap on embedding-based source retrieval.} This figure shows how the number of overlapping words between target and source names affects retrieval performance, measured by the percentage of times the gold source is retrieved and its average rank (lower is better). As lexical overlap increases, both retrieval success and ranking quality improve, indicating that embedding-based retrieval is strongly influenced by surface-level word overlap.}
    \label{fig:wordoverlap}
\end{figure}

Figure \ref{fig:wordoverlap} shows the impact of lexical overlap on embedding-based source retrieval.
\label{app:wordoverlapp}

\end{document}